\documentclass[12pt]{article}
\usepackage[utf8]{inputenc}
\usepackage{amsmath,amssymb}
\usepackage{graphicx,epsf, color}
\usepackage{enumerate}
\usepackage{natbib}
\usepackage{forest}
\usepackage{float}
\usepackage{url} % not crucial - just used below for the URL
\usepackage{times}
\usepackage[ruled,noend]{algorithm2e}
\usepackage{graphics}
\usepackage{booktabs}
\usepackage[colorlinks=true, linkcolor=blue, citecolor=blue, urlcolor = blue, filecolor=blue]{hyperref}
\usepackage{setspace}
\usepackage{caption}
\usepackage{subcaption}
\usepackage{lineno}

\newcommand{\blind}{0}

\usepackage[margin=1in]{geometry}

\catcode`\^ = 13 \def^#1{\sp{#1}{}}
\definecolor{Pink}{rgb}{1.0, 0.5, 0.5}
\definecolor{Maroon}{rgb}{0.8, 0.0, 0.0}

\def\boxit#1{\vbox{\hrule\hbox{\vrule\kern6pt\vbox{\kern6pt#1\kern6pt}\kern6pt\vrule}\hrule}}

\newcommand{\M}{\mbox}

\newtheorem{theorem}{Theorem}[section]
\newtheorem{lemma}[theorem]{Lemma}

\newtheorem{proposition}[theorem]{Proposition}

\newcommand{\bone}{\mbox{\bf 1}}
\newcommand{\bzero}{\mbox{\bf 0}}

\newcommand{\bdelta}{\mbox{\boldmath $\delta$}}

\newcommand{\bbeta}{\mbox{\boldmath $\beta$}}

\newcommand{\bgamma}{\mbox{\boldmath $\gamma$}}

\newcommand{\var}{\mathrm{var}}

\newcommand{\diag}{\mathrm{diag}}

\newcommand{\rank}{\mathrm{rank}}
\newcommand{\sgn}{\mathrm{sgn}}

\def\t{^\top}

\def\beqn{\begin{eqnarray}}
\def\eeqn{\end{eqnarray}}
\def\beqns{\begin{eqnarray*}}
\def\eeqns{\end{eqnarray*}}

\def\0{{\bf 0}}
\def\A{{\bf A}}

\def\b{{\bf b}}

\def\D{{\bf D}}

\def\I{{\bf I}}

\def\M{{\bf M}}

\def\t{{\bf t}}

\def\V{{\bf V}}

\def\s{{\bf s}}
\def\u{{\bf u}}
\def\v{{\bf v}}

\def\X{{\bf X}}
\def\x{{\bf x}}

\def\y{{\bf y}}

\def\z{{\bf z}}
\def\1{{\bf 1}}

\def\trans{^{\rm T}}

\def\t1trans{^{t+1\rm T}}

\begin{document}

\def\spacingset#1{\renewcommand{\baselinestretch}%
{#1}\small\normalsize}
\spacingset{1}

%\pagenumbering{roman}
%%%%%%%%%%%%%%%%%%%%%%%%%%%%%%%%%%%%%%%%%%%%%%%%%%%%%%%%%%%%%%%%%%%%%%%%%%%%%%
\title{\bf Majorization-Minimization Dual Stagewise Algorithm for Generalized Lasso}

% \if1\blind
% {  \author{Jianmin Chen, Kun Chen}

% } \fi

\if1\blind
{
  \author{}\date{}
  \maketitle
} \fi

\if0\blind
{
  \author{Jianmin Chen$^1$, Kun Chen$^{2}$\thanks{Corresponding author; kun.chen@uconn.edu}\\    
$^1$\textit{Department of Department of Biostatistics, Epidemiology and Informatics,}\\ \textit{University of Pennsylvania}\\%[2pt]
$^2$\textit{Department of Statistics, University of Connecticut}
}
\maketitle
}
\fi

%\date{}

%\maketitle

% \bigskip

\begin{abstract}

The generalized lasso is a natural generalization of the celebrated lasso approach to handle structural regularization problems. Many important methods and applications fall into this framework, including fused lasso, clustered lasso, and constrained lasso. To elevate its effectiveness in large-scale problems, extensive research has been conducted on the computational strategies of generalized lasso. However, to our knowledge, most studies are under the linear setup, with limited advances in non-Gaussian and non-linear models. We propose a \textbf{m}ajorization-\textbf{m}inimization \textbf{du}al \textbf{st}agewise (MM-DUST) algorithm to efficiently trace out the full solution paths of the generalized lasso problem. The majorization technique is incorporated to handle different convex loss functions through their quadratic majorizers. Utilizing the connection between primal and dual problems and the idea of ``slow-brewing'' from stagewise learning, the minimization step is carried out in the dual space through a sequence of simple coordinate-wise updates on the dual coefficients with a small step size. Consequently, selecting an appropriate step size enables a trade-off between statistical accuracy and computational efficiency. We analyze the computational complexity of MM-DUST and establish the uniform convergence of the approximated solution paths. Extensive simulation studies and applications with regularized logistic regression and Cox model demonstrate the effectiveness of the proposed approach. 
\end{abstract}

\noindent%

% \vfill
\doublespacing
%\spacingset{1.45} % DON'T change the spacing! % 25 lines per page

%%%%%%%%%%%%%%%%%%%%%%%%%%%%%%%%%%%%%%%%%%%%%%
%%
%% Start From Here
%%
%%%%%%%%%%%%%%%%%%%%%%%%%%%%%%%%%%%%%%%%%%%%%%%
% \pagenumbering{arabic}
% \setcounter{page}{1}

\clearpage

%\linenumbers

\section{Introduction}

Regularization techniques are fundamental in modern statistics and machine learning. Arguably, one of the most popular regularization methods is the \emph{lasso} method \citep{buhlmann2011statistics}, while the generalized lasso~\citep{tibshirani2011solution} method is one of its natural generalizations. Instead of penalizing the $\ell_1$ norm of a parameter vector, say, $\bbeta \in \mathbb{R}^p$, the generalized lasso applies the $\ell_1$ penalty on a linear transformation of $\bbeta$ to enable structural regularization. Without loss of generality, we consider the following generalized lasso formulation,
\begin{align}
    \mathop{\mbox{minimize}}\limits_{\bbeta}\ f(\bbeta) + \lambda \|\D\bbeta\|_1, \label{eq:optim-glasso}
\end{align}
where $\lambda\geq 0$ is the tuning parameter, $\bbeta\in\mathbb{R}^p$ is the coefficient vector, $f(\cdot)$ is a convex and differentiable loss function, and $\D\in\mathbb{R}^{m\times p}$ is the structural matrix defined based on the specific problem of interest.

The flexibility of specifying $\D$ enables broad applications, and several well-known regularized estimation problems can be cast as special cases within this general framework in \eqref{eq:optim-glasso}. For example, the fused lasso and trend-filtering are widely applied for signal approximation or denoising with time series and image data~\citep{wang2015trend,arnold2016efficient}, in which the matrix $\D$ is usually designed to enforce sparsity on the differences between neighboring data points or parameters. More broadly, different variations of the generalized lasso can be represented with different $\D$ matrices to capture structural patterns among features or parameters. 
%More broadly, the customizable $\D$ matrix can be used to capture the structural patterns among features or parameters. 
In the most trivial case, when $\D$ is the identity matrix, the generalized lasso reduces to a lasso problem. Several non-trivial scenarios were discussed in \citet{tibshirani2011solution}, where the transformation to lasso becomes impossible when $\D$ is not of full row-rank. 
Examples include the clustered lasso \citep{she2010sparse}, the spatially varying coefficients regression \citep{zhao2020solution}, the relative-shift regression \citep{li2022pursuing} in which the $\D$ matrix is designed to allocate heterogeneity effects among features, and the feature selection and aggregation problems \citep{yan2021rare} where $\D$ is constructed from a hierarchical structure among the features. Further, constrained lasso is also strongly connected with generalized lasso and can be converted to it under certain conditions  \citep{gaines2018algorithms}. 

%{\color{red} This paragraph could be removed or moved to a later section. This feels like a distraction.}{\color{blue} I agree this is a distraction. I replaced this part with the one sentence above.}

%Interestingly, the generalized lasso is strongly connected to the constrained lasso problem. Specifically, when $\D$ is of full column-rank, we can always find a matrix $\D^+\in\mathbb{R}^{p\times m}$, such that $\D^+\D=\I_p$. 
%By introducing a transformation where $\D\bbeta=\z$ and $\bbeta=\D^+\z$, the problem in~\eqref{eq:optim-glasso} can be rewritten as 
%\begin{align*}
%    \mathop{\mbox{minimize}}\limits_{\bz}\ f(\D^+\z)+\lambda\|\z\|_1, \mbox{ s.t. } \z \in \mbox{col}(\D).
%\end{align*}
%Indeed, any generalized lasso problem can be converted to a constrained lasso \citep{gaines2018algorithms}. On the other hand, a constrained lasso problem can be converted to the generalized lasso framework when the constraints on $\bbeta$ can be expressed as $\C\bbeta=\bzero$, with $\C$ being a matrix of full row rank. {\color{red} This paragraph could be removed or moved to a later section. This feels like a distraction.}

With various important applications, extensive research has been conducted on the optimization methods for the generalized lasso. It is worth noting that, in general, its optimization is more complicated than the lasso problem, as the $\D$ matrix makes the penalty part non-separable. The existing studies and discussions have mainly concentrated on the case when the loss $f(\cdot)$ is a quadratic function in $\bbeta$. While \citet{hoefling2010path} and \citet{arnold2016efficient} designed the \emph{LARS} path-following algorithms for low-dimensional cases, the proximal gradient descent (PDG) and alternating direction method of multipliers (ADMM) were shown to be more efficient when dealing with high-dimensional problems~\citep{wang2015trend,chung2022variable, zhong2023sparse}. There are many other similar algorithms, including those proposed in \citet{liu2010efficient}, \citet{she2010sparse}, and \citet{gaines2018algorithms}.

%The existing studies and discussions have mainly concentrated on the case when the loss $f(\cdot)$ is a quadratic function in $\bbeta$.%, with limited exploration for general loss functions. 

%The wide range of applications for generalized lasso has prompted extensive research into the optimization methods. Unlike the standard lasso, the generalized lasso introduces additional complexity as the $\D$ matrix makes the penalty function non-separable. Most existing studies focus on cases where the loss function $f(\cdot)$ is quadratic in $\bbeta$, which simplifies the optimization. For low-dimensional settings, \citet{hoefling2010path} and \citet{arnold2016efficient} developed path-following algorithms based on the \emph{LARS} framework. In contrast, for high-dimensional problems, proximal gradient descent (PGD) and alternating direction method of multipliers (ADMM) have demonstrated greater computational efficiency \citep{wang2015trend, chung2022variable, zhong2023sparse}. Other notable methods include those proposed by \citet{liu2010efficient}, \citet{she2010sparse}, \citet{lin2014alternating}, and \citet{gaines2018algorithms}.

Optimization techniques for the generalized lasso with general loss functions remain underexplored. Efficient algorithms for the fused lasso under generalized linear models have been developed \citep{ye2011split, xin2014efficient, tang2016fused}, but their extension to a more general $\D$ matrix can be highly non-trivial. \citet{zhou2014generic} proposed a path-following method applicable to general loss functions and $\D$ matrices; however, the method requires solving an ordinary differential equation, which is not desired for large-scale or high-dimensional problems. Alternative approaches, such as \citet{chen2012smoothing}, \citet{lin2014alternating}, and \citet{zhu2017augmented}, offer a wider applicability for generalized lasso, but remain computationally expensive due to the complexity of gradient updates for general loss functions.

We propose a novel {\bf m}ajorization-{\bf m}inimization {\bf du}al {\bf st}agewise (MM-DUST) algorithm to efficiently trace out the solution paths of the generalized lasso problem with a general loss function. Our algorithm utilizes three key ideas. First, the main architecture of our algorithm is based on the \emph{majorization-minimization} (MM) technique~\citep{hunter2004tutorial}, which is used to handle the general loss function ~\citep{schifano2010majorization, yang2012cmd, jiang2014majorization, yu2015high}. Specifically, we iteratively alternate between a majorization step and a minimization step; with some initial values or the current estimates, the majorization step constructs a surrogate objective function, i.e., a majorizer of the original objective function, that becomes much easier to optimize in the subsequent minimization step. %that is much easier to optimize than the original objective function. The surrogate function is then minimized in the minimization step. %The surrogate is constructed for the penalty function in \citet{schifano2010majorization} and  \citet{yu2015high}, while \citet{yang2012cmd} and \citet{jiang2014majorization} try to construct the quadratic approximation on the loss function. 
In our work, we take the quadratic approximation of the complicated loss function (when it is convex) in the majorization step to result in a regular generalized lasso problem with squared error loss. %in order to obtain a simpler form for the dual problem of~\eqref{eq:optim-glasso}. %and get rid of the non-separable penalty function. 
Second, instead of directly solving the generalized lasso problem, we target its dual problem, which becomes a box-constrained convex problem. Finally, the minimization is conducted by solving the problem from the dual space with a stagewise descent strategy. We borrow the ``slow-brewing'' idea from \emph{stagewise} learning, where the model complexity gradually grows through a sequence of simple updates, and potential computational benefits can be obtained by properly selecting the ``learning rate'' in each simple update~\citep{zhao2007stagewise,tibshirani2015general,tibshirani2011solution, chen2022fast}. Instead of fully solving the optimization problem in each minimization step, we only update the dual parameters with a limited number of iterations. In each iteration, the dual parameters are changed by a small step size along the directions that lead to the greatest descent of the dual objective. Our algorithm then proceeds to generate approximated solution paths with increasing model complexity until an early stopping criterion is met and a desired solution is reached.

Our work is among the first to provide a path algorithm for a generalized lasso problem with a general loss function. The choice of step size in the minimization step provides a venue to balance computational efficiency and statistical accuracy. Theoretical analysis provides a strong uniform convergence guarantee and potential computational complexity reduction, and extensive numerical studies further demonstrate the effectiveness and scalability of our approach. 
%by novelty combining the majorization-minimization technique with the ``slow-learning'' idea. 

The rest of the paper is organized as follows. Section~\ref{sec:dual} presents the dual problem of generalized lasso and gives a new algorithm to solve the dual problem. We introduce the MM-DUST algorithm and discuss its theoretical properties in Section~\ref{sec:glasso-MM-DUST}. In Section~\ref{sec:glasso-simu}, we conduct simulation studies to demonstrate the convergence of the path algorithm and explore its accuracy and efficiency. Two real data applications are presented in Section~\ref{sec:glasso-real} with further discussions in Section~\ref{sec:glasso-dis}.
\section{Primal \& Dual Problems of Generalized Lasso}\label{sec:dual}

While the generalized lasso problem in ~\eqref{eq:optim-glasso} has a unique solution when the loss function $f(\cdot)$ is strictly convex, direct optimization can be complicated as the penalty function is non-separable and non-smooth with a general $\D$ matrix. A common strategy in optimization is to consider the dual problem. In this section, we explore the duality of the generalized lasso and provide an algorithm to solve it through its dual.  

\subsection{Dual Problem}\label{sec:glasso-dual}

To study the duality, we first augment the original problem in~\eqref{eq:optim-glasso} by introducing a new vector $\z\in\mathbb{R}^{m}$ as the following
\begin{align}
    \mathop{\mbox{minimize}}\limits_{\bbeta, \z}\ f(\bbeta) + \lambda \|\z\|_1, \mbox{ s.t. } \z = \D\bbeta.
    \label{eq:initial-prob-aug}
\end{align}
Its Lagrangian form is then 
\begin{align}
    \mathop{\mbox{minimize}}\limits_{\bbeta, \z, \u}\ f(\bbeta) + \lambda \|\z\|_1 + \u^{\trans}(\D\bbeta-\z),
    \label{eq:primal1}
\end{align}
where $\u\in\mathbb{R}^m$ is the Lagrangian multiplier. We adopt the standard terminologies from convex optimization and 
consider problem~\eqref{eq:primal1} as the primal problem. It follows that the dual problem of \eqref{eq:primal1} has the following form
\begin{align}
    \mathop{\mbox{minimize}}\limits_{\u}\ f^*(-\D^{\trans}\u), \mbox{ s.t. } \|\u\|_{\infty} \leq \lambda,
    \label{eq:dual}
\end{align}
where $f^*(\cdot)$ is the convex conjugate function of $f(\cdot)$; detailed derivation for the dual problem is provided in Appendix~\ref{supp:dual}. %Instead of solving the complicated primal problem, 
We can solve the dual problem and get the solution of the primal variable $\bbeta$ through the primal-dual stationarity condition at the optimal point
\begin{align}
    \nabla f(\bbeta)\ +\ \D^{\trans}\u\ =\ \0, \label{eq:primal-dual}
\end{align}
where $\nabla$ is the gradient operator.

It is worth noting that a large $\lambda$ value suggests heavy regularization on the primal problem but little restriction on 
the dual problem, while a small $\lambda$ value indicates the reversed scenario. Thus, it is natural to build the 
solution paths starting at an empty primal model but a full dual model along a decreasing sequence of $\lambda$ values.

\subsection{Dual Algorithm}\label{sec:glassoalg}

Since the conjugate function $f^*(\cdot)$ is always convex, we can see that the dual problem is 
actually a \emph{box-constrained} problem with a convex objective function $f^*(\cdot)$. It can be readily proved that a convex box-constrained problem can be solved by a coordinate-wise algorithm with suboptimality as stated in Lemma~\ref{lemma:dual}; see detailed proof in Appendix~\ref{app:glasso-proof}.  

%\begin{lemma}\label{lemma:dual}
%For a box-constrained problem with convex and differentiable objective function, if we can find a coordinate-wise minimum point such that the 
%objective function value cannot be further reduced in any feasible direction, then this point is the optimal solution to the problem.
%\end{lemma}

\begin{lemma}\label{lemma:dual}
Consider a box-constrained problem with a loss function $f(\cdot)$, which is convex, differentiable and $L$-smooth. With boundary value $\lambda$ in the constraint, consider a point $\hat\u$ such that
\begin{align*}
\begin{cases}
    f(\hat\u \pm \varepsilon\bone_j) \geq f(\hat\u), & |\hat u_j|\leq \lambda,\\
    f(\hat\u - \varepsilon\bone_j\sgn(u_j)) \geq f(\hat\u), & |\hat u_j|= \lambda.\\
\end{cases}  
\end{align*}
Then $f(\hat\u)$ converges to $f(\u^*)$ when $\varepsilon\rightarrow 0$, and $\u^*$ is a minimizer of the box-constrained problem.
\end{lemma}
% We remark that the suboptimality result does not require strong convexity of the loss function. 

Based on Lemma~\ref{lemma:dual}, we design a stagewise descent algorithm to solve the dual problem~\eqref{eq:dual} with a non-decreasing $\lambda$ sequence over the iterations. Different from coordinate descent which updates each dimension in a cyclic order, the stagewise approach selects the direction that leads to the greatest decrease in the loss function, and then takes only a small step in the chosen direction. 

Without loss of generality, let $f(\cdot)$ represent the loss function in a box-constrained problem. Denote $\u^F$ as the solution to the dual problem when $\lambda = \infty$, where there is no constraint imposed on $\u$. The details are stated in Algorithm~\ref{alg:dual}.

\begin{algorithm}[tbp]
    \caption{Stagewise Descent Algorithm for Box-Constrained Problem}
    \label{alg:dual}
    \KwIn{$\X$, $\y$, $\u^F$, a small step size constant $\varepsilon>0$. }
    \textbf{Step 1: Initialization}. Round $\u_F$ by $\varepsilon$. Then take an initial backward step
    \begin{align*}
        \mathcal{K}_0\ &=\ \mathop{\arg\max}\limits_{1\leq i \leq m}|u_{i}^F|, \\
        \u_0\ &=\ \u^F\ -\ \varepsilon \sgn(\u^F) \circ \bone_{\mathcal{K}_0}, \\
        \lambda_0\ &=\ \max_{1\leq i \leq m}|u_{0i}|, 
    \end{align*}
    where $\circ$ stands for element-wise multiplication, and $\bone_{\mathcal{K}_0}$ is a $m$-dimensional vector with the $i$th entry to be $1$ for all $i\in\mathcal{K}_0$ and $0$ otherwise.\linebreak
    
    \textbf{Step 2: Stagewise descent at fixed $\lambda_t$ value.}
    At the $t$th iteration, check if there exists a feasible descent direction. Find the steepest descent direction on the loss function $f$ among all the feasible directions: 
    \begin{align*}
        (\hat i, \ \hat s_{\hat i})\ =\ \mathop{\arg\min}\limits_{s=\pm\varepsilon} f(\u_t+s\bone_i),\ \mbox{s.t.}\ 
        \|\u_t\ +\ s\bone_i\|_{\infty}\ \leq\ \lambda_t.
    \end{align*}
    If $f(\u_t+\hat{s}_{\hat i}\bone_{\hat i})-f(\u_t)\ <\ 0$, then take the update: 
    \begin{align*}
        \u_{t+1}\ =\ \u_t\ +\ \hat{s}_{\hat i}\bone_{\hat i}.
    \end{align*}
    Otherwise, force a backward step and decrease $\lambda$,
    \begin{align*}
        \mathcal{K}_t\ &=\ \mathop{\arg\max}\limits_{1\leq i \leq m}|\u_{ti}|, \\
        \u_{t+1}\ &=\ \u_t\ -\ \varepsilon \sgn(\u_t) \circ \bone_{\mathcal{K}_t}, \\
        \lambda_{t+1}\ &=\ \lambda_t\ -\ \varepsilon.
    \end{align*}
    
    \textbf{Iteration}. Increase $t$ by one and repeat step $2$ until 
    $\lambda_t \leq \varepsilon$.
\end{algorithm}

The convergence of Algorithm~\ref{alg:dual} is guaranteed by Lemma~\ref{lemma:dual}. Note that the parameter $\lambda$ serves as a ``boundary'' for the dual parameter $\u$, and any descent step will not happen in the direction that widens the boundary. Additionally, the step size will determine the number of $\lambda$ values in the sequence and the number of points on the full solution paths. For every $\u_t$ at $\lambda_t$ along the dual solution paths, we can find a corresponding primal vector $\bbeta_t$ by solving the primal-dual stationarity condition specified in ~\eqref{eq:primal-dual}. %{\color{red} How efficient is this? Comment?} %As a result, we can derive the primal solution paths from the dual solution paths.

%{\color{blue} I tried to summarize the information in Section 2.3 here, as the discussion in Section~2.3 is not the major work for this paper. If okay, we can delete Section 2.3}

As an example, consider the dual problem~\eqref{eq:dual} under the special case of a squared error loss function. Assume the primal loss is $f(\bbeta)=1/2\|\y-\X\bbeta\|^2$, with $\X\in\mathbb{R}^{n\times p}$ and $r(\X)\leq p$. It turns out that the explicit form of the dual can be derived with a simple quadratic loss as the following
\begin{align*}
    \mathop{\mbox{minimize}}\limits_{\u}f^*(-\D^{\trans}\u), \mbox{ s.t. } \D^{\trans}\u\in\mbox{row}(\X),\  \mbox{and }\|\u\|_{\infty}\leq\lambda,
\end{align*}
where
\begin{align*}
    f^*(-\D^{\trans}\u)\ =\ 1/2(\X^{\trans}\y-\D^{\trans}\u)^{\trans}(\X^{\trans}\X)^+(\X^{\trans}\y-\D^{\trans}\u),
\end{align*}
and $(\X^{\trans}\X)^+$ is the Moore-Penrose inverse of $\X^{\trans}\X$. When $r(\X)=p$, the extra constraint $\D^{\trans}\u\in\mbox{row}(\X)$ is automatically satisfied. When $r(\X)<p$, as suggested by \citet{tibshirani2011solution}, we can add a small $\ell_2$ penalty on $\bbeta$ in the original primal problem to remove the constraint. It is then straightforward to implement Algorithm~\ref{alg:dual} to solve a quadratic generalized lasso problem.

With the simple squared loss, another similar dual algorithm was discussed in \citet{tibshirani2011solution}, and the theoretical properties of the solutions have also been studied \citep{tibshirani2012degrees, arnold2016efficient}. Nonetheless, adapting these algorithms to more complex loss functions is non-trivial and remains an open problem.

\section{Primal-Dual Iterations with Majorization \& Stagewise Learning}

\label{sec:glasso-MM-DUST}

\subsection{Motivating Examples}

When more complicated loss functions are considered in the generalized lasso problem, finding the conjugate function can be challenging, or the conjugate function does not have an explicit form, rendering  Algorithm \ref{alg:dual} in Section \ref{sec:dual} ``useless".

Specifically, we focus on two widely-used statistical models in this work: 
\begin{itemize}
    \item Logistic regression: Let $\y\in\mathbb{R}^n$ be the binary response vector. The loss function is the negative log-likelihood, where
    \begin{align*}
            f(\bbeta) = \sum_{i=1}^n -y_i\x_i^{\trans}\bbeta+\log (1+\exp(\x_i^{\trans}\bbeta)).
    \end{align*}
    \item Cox regression: Let $\mathcal{D}$ be the index set of subjects that have the observed event, and $\mathcal{R}_s$ be the risk set of the $s$th event in the data set. We consider the negative log partial likelihood when there are no ties as the loss function, where
    \begin{align*}
            f(\bbeta) = \sum_{s\in\mathcal{D}}(-\x_s^{\trans}\bbeta + \log\sum_{i\in\mathcal{R}_s}\exp(\x_i^{\trans}\bbeta)).
    \end{align*}
\end{itemize}
Generalized lasso have been combined with these two regression models with broad applications including but not limited to spectral data with temporal structure~\citep{yu2015classification}, brain data with spatial structure~\citep{lee2014application, xin2014efficient}, genomic data with location information~\citep{chaturvedi2014fused}, and network data~\citep{tran2020classifying}.

%{\color{red} Add a few sentences and references to demonstrate that there are many generalized lasso problems associated with binary classification and survival analysis.}{\color{blue} Added.}

In either case, the loss function is convex, but the explicit form of the conjugate function is intractable. Besides, calculating the gradient of the loss function can be a time-consuming process, yet it is required in every iteration step by most of the algorithms. % as well as Algorithm~\ref{alg:dual}. 
Furthermore, to get the primal solution, we need to evaluate the primal-dual stationarity condition for each pair of $\bbeta_t$ and $\u_t$. When equation~\eqref{eq:primal-dual} cannot be simplified, solving the primal parameter from the dual parameter amounts to dealing with a ``root-finding'' problem, which can also be expensive in computation.

\subsection{Proposed Algorithm}

We develop a majorization-minimization dual stagewise (MM-DUST) algorithm to appriximately trace out the
full solution paths of the generalized lasso problem. 

First, to solve the generalized lasso problem from the dual, we incorporate the idea of ``majorization-minimization'' with the primal-dual iterations. Instead of solving the initial primal problem with a complicated loss function, in the majorization step, we approximate the original loss function with a quadratic majorizer and optimize the approximated problem in the minimization step. 

Assume that the primal loss function $f(\cdot)$ is convex and twice continuously differentiable. Suppose we can find a matrix $\M\in\mathbb{R}^{p\times p}$, such that $\M-\nabla^2f(\bbeta)$ is positive semi-definite for all $\bbeta\in\mathbb{R}^p$, then a quadratic majorizer of $f(\cdot)$ at an initial point $\bbeta_0\in\mathbb{R}^p$ can be taken as 
\begin{align*}
    \tilde f(\bbeta|\bbeta_0) =  f(\bbeta^0)+\nabla f(\bbeta_0)^{\trans}(\bbeta-\bbeta_0)+\frac{1}{2}(\bbeta-\bbeta_0)^{\trans}\M(\bbeta-\bbeta_0).
\end{align*}
We then take $\tilde f(\bbeta|\bbeta_0)$ as the loss function in the primal problem~\eqref{eq:primal} with a fixed tuning parameter $\lambda$. 
Further, when $\M$ is a diagonal matrix and $\M$ can be expressed as $\M=L\I_p$ with $L$ being a positive constant, the primal minimization problem can be simplified as
\begin{align}
    \mathop{\mbox{minimize}}\limits_{\bbeta,\z,\u}\ \frac{1}{2L}\|\widetilde{\y}-L\bbeta\|^2 + \lambda \|\z\|_1 + \u^{\trans}(\D\bbeta-\z),\ \widetilde{\y}=L\bbeta_0-\nabla f(\bbeta_0).
    \label{eq:primalmm}
\end{align}
For the logistic loss, it is well known that the constant $L$ can be taken as $\sigma_m/4$, where $\sigma_m$ is the largest eigen value of $\X^{\trans}\X$ with $\X$ being the primal design matrix. For the Cox model, the constant $L$ is calculated based on our derivations in Appendix~\ref{sec:coxproof}.

After the quadratic approximation, the primal problem is in the form of a least square problem with a full rank design matrix $L\I_p$. Building on the derivations of the dual problem in Section~\ref{sec:glassoalg}, the corresponding dual problem can be formulated as 
\begin{align}
    \mathop{\mbox{minimize}}\limits_{\u}\ \frac{1}{2L}\|\widetilde{\y}-\D^{\trans}\u\|^2,\ \widetilde{\y}=L\bbeta_0-\nabla f(\bbeta_0),\ 
    \mbox{s.t. } \|\u\|_{\infty}\leq\lambda.
    \label{eq:dualmm}
\end{align}
When problem~\eqref{eq:dualmm} is optimized at $\hat\u$, the primal estimate $\hat\bbeta$ can be derived from the primal-dual stationarity condition \eqref{eq:primal-dual}, which can be simplified to the following updating rule
\begin{align}
    \hat\bbeta\ =\ \bbeta_0\ -\ \frac{1}{L}(\D^{\trans}\hat\u+\nabla f(\bbeta_0)).
    \label{eq:primal-dual2}
\end{align}
With the discussions above, we consider the optimization strategy as alternating between taking the majorizer as in \eqref{eq:primalmm}, solving the dual problem as in \eqref{eq:dualmm}, and updating the primal estimate as in \eqref{eq:primal-dual2}.

\begin{figure}[H]
\centering
\includegraphics[width=0.9\textwidth]{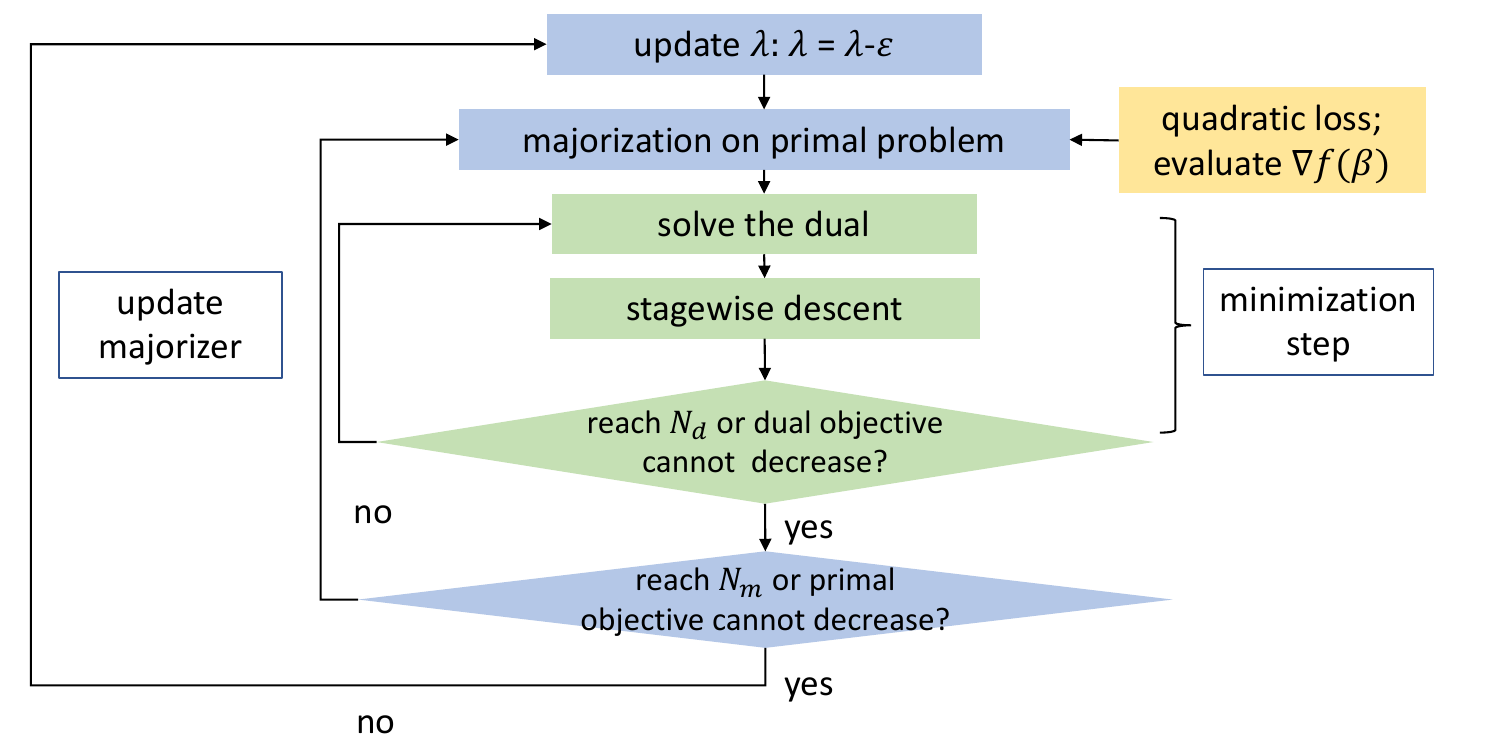}
\caption{Overview of the MM-DUST algorithm.}\label{fig:MM-DUST-flow}
\end{figure}

Further, to obtain the advantages of ``slow-brewing'', we combine the primal-dual iterations with the stagewise descent Algorithm~\ref{alg:dual}. The updating flow is illustrated in Figure~\ref{fig:MM-DUST-flow}. 

Starting with the first majorization step on the primal at current $\lambda$, we transition to the dual space and update $\u$ for several iterations with a small step size $\varepsilon$. The stagewise descend rule is applied for the dual updating.  
Instead of taking the backward step and reducing $\lambda$ when there is no descent direction, we terminate the dual updates after a limited number of simple iterations. At the same time, $\bbeta$ can be updated and the next majorizer at the current $\lambda$ is measured. 

%we take the backward step after a limited number of simple iterations. At the same time, $\bbeta$ can be updated and the next majorizer at the current $\lambda$ is measured. 

We ensure the minimization step by monitoring the change in the initial primal objective function. However, in practice, updating the dual will usually lead to a decrement in the primal problem. Once the initial primal objective function cannot be further reduced, $\lambda$ is decreased by $\varepsilon$ and we take the backward step on $\u$. The updating  details are summarized in Algorithm~\ref{alg:MM-DUST} and Algorithm~\ref{alg:dualsolver}. We use $N_m$, $N_d$ to denote the maximum number of majorization steps for each $\lambda$, and the maximum number of iterations in each dual problem, respectively.

\begin{algorithm}[tbp]
	\caption{Majorization-Minimization Dual Stagewise Algorithm (MM-DUST)}
    \label{alg:MM-DUST}
        \KwIn{loss function $f$, $\X$, $\y$, $\D$, $N_m$, $N_d$, $\varepsilon$}
        \textbf{Initialization:} 
        
        1. Calculate constant $L$ based on loss function $f$, $\X$ and $\y$\;
        2. Solve initial values $\hat\bbeta_0$ and $\hat\u_0$ from \\
        $$\hat\bbeta_0 = \arg\min f(\bbeta),\,\mbox{s.t. }\|\D\bbeta\|_1=\bzero,$$ 
        $$\hat\u_0 = \arg\min \frac{1}{2L}\|L\hat\bbeta_0 - \nabla f(\hat\bbeta_0) - \D^{\trans}\u\|;$$
        
        3. Round $\hat\u_0$ by $\varepsilon$ \;
        4. Get $\lambda_0 = \max_{1\leq i\leq m}|\hat u_{0i}|$, $N_{\lambda} = \lambda_0/\varepsilon$, and $\Gamma(\lambda_0)=f(\hat\bbeta_0)+\lambda_0\|\D\hat\bbeta_0\|_1$.

        \For{$1\leq t < N_{\lambda}$}{
        Take an initial backward step on $\lambda$ and $\u$: \\
        %Take an initial backward step on $\u$: 
        %$\mathcal{K} = \arg\max_{1\leq i \leq m}|\hat u_{(t-1)i}|,\ 
        %\u = \u_0 - \varepsilon \sgn(\u_0)\circ\bone_{\mathcal{K}}$
        $\lambda_t=\lambda_{t-1}-\varepsilon$, $\mathcal{K} = \arg\max_{1\leq i \leq m}|\hat u_{(t-1)i}|,\ 
        \hat\u_t = \hat\u_{t-1} - \varepsilon \sgn(\hat\u_{t-1})\circ\bone_{\mathcal{K}}$ \;
       % Primal update: $\bbeta(\lambda^t) = \bbeta(\lambda^{t-1})-\frac{1}{L}[\D^{\trans}u(\lambda^t)+\nabla f(\bbeta(\lambda^{t-1}))]$ \;
        Surrogate update: $\tilde \y_t = L\hat\bbeta_{t-1} - \nabla f(\hat\bbeta_{t-1})$ \;
        Set: $\hat\bbeta_t = \hat\bbeta_{t-1}$, $\Gamma(\lambda_t) = f(\hat\bbeta_t)+\lambda_t\|\D\hat\bbeta_{t}\|_1$.
        \linebreak
        \For{$1\leq j \leq N_m$}{
        1. $\u_{inner}^j = \mbox{\textbf{Dual-Solver}}
        (\tilde \y_t, \D, \hat\u_{t}, N_d, \varepsilon)$\;
        2. $\bbeta_{inner}^j = \hat\bbeta_t - \frac{1}{L}[\D^{\trans}\u_{inner}^j+\nabla f(\hat\bbeta_t)]$  \;
        3. $\Gamma_{inner}^j = f(\bbeta_{inner}^j)+\lambda_t\|\D\bbeta_{inner}^j\|_1$\;
        4. Update surrogate function with the same $\lambda$ value only when the objective function $\Gamma$ is decreasing. \linebreak
        \eIf{($\Gamma_{inner}^j > \Gamma(\lambda_t)$)}{
        break\;
        }
        {$\hat\u_t = \u_{inner}^j$, $\hat\bbeta_t = \bbeta_{inner}^j$, \\
        $\tilde \y_t = L\hat\bbeta_t - \nabla f(\hat\bbeta_t),\ \Gamma(\lambda_t)=\Gamma_{inner}^j$.
        }
        \textbf{end};
        }
        \textbf{end} \linebreak
        
    }
    \textbf{end}
    
    \KwOut{All values of $\u$ and $\bbeta$ with a decreasing sequence of $\lambda$.}
\end{algorithm}

\begin{algorithm}[tbp]
	\caption{Dual-Solver}
    \label{alg:dualsolver}
        \KwIn{Dual response vector $\tilde{\y}$, dual design matrix $\D^{\trans}$, dual initial value $\u_0$, $N_d$, $\varepsilon$.} 
        %Take an initial backward step on $\u_0$: 
        %$\mathcal{K} = \arg\max_{1\leq i \leq m}|u_{0i}|,\ 
        %\u = \u_0 - \varepsilon \sgn(\u_0)\circ\bone_{\mathcal{K}}$
	$c = \max_{1\leq i\leq m}|u_{0i}|$.

        \For{$1\leq t \leq N_{d}$}{
        $(\hat i, \ \hat s_{\hat i})\ =\ \mathop{\arg\min}\limits_{s=\pm\varepsilon}\|\tilde\y - \D^{\trans}(\u+s\bone_i)\|^2,\ \mbox{s.t.}\ 
        \|\u+s\bone_i\|_{\infty}\ \leq\ c.$

        \eIf{$\|\tilde\y - \D^{\trans}(\u+\hat{s}\bone_{\hat i})\|^2 < \|\tilde\y - \D^{\trans}\u\|^2$}{
        take the update: $\u\ =\ \u\ +\ \hat{s}_{\hat i}\bone_{\hat i}$ \;
        }
        { 
        break.
        }
        \textbf{end}
    }
    \textbf{end}
    
    \KwOut{$\u$ at the last iteration.}
\end{algorithm}

\subsection{Algorithmic Convergence \& Complexity}
\label{sec:comp}

We present the algorithmic convergence and complexity of the proposed MM-DUST approach. All the detailed derivations are provided in Appendix~\ref{app:sec:conv}.

Under certain conditions, we show that the solution paths provided by MM-DUST converge to the exact solution paths of a generalized lasso problem, which is stated in the next theorem.
Recall that $N_m$ and $N_d$ denote the maximum number of majorization steps for each $\lambda$ value and the maximum number of iterations in each dual problem, respectively.

\begin{theorem}\label{thm:thm1}
Let $N_0$ be the minimum number of iterations required for all the dual problems to reach a point where no descent direction exists for an $\varepsilon$ step size.
If the loss function $f(\bbeta)$ is $\mu$-strongly convex and $L$-smooth, then as the step size $\varepsilon \rightarrow 0$, $|N_0-N_d|=O(1)$ and when we take $N_m=1$, the MM-DUST paths converge to the generalized lasso paths uniformly.
\end{theorem}
%There are several important parameters in Algorithm~\ref{alg:MM-DUST}. 
Intuitively, $\varepsilon$ controls the number of majorizers we need to compute along the entire solution paths, as well as the number of $\lambda$ values. It is worth noting that the time-consuming evaluation of the first-order gradient of the initial primal loss only occurs at the majorization step, and the number of such steps is controlled by $\varepsilon$ and $N_m$. On the other hand, having more $\lambda$ values allows us to approach the exact solution paths more accurately. As a result, the magnitude of $\varepsilon$ strikes a balance between the path-fitting accuracy and the computational efficiency. In addition to it, the number of dual iterations, $N_d$, affects the optimality of the dual solution. Insufficient iterations in the dual may violate the primal-dual stationarity condition derived at the optimal points of both the dual and primal problems. 

%{\color{red} Can you add something about the computation complexity? You can check my stagewise learning paper for a similar discussion; it would be nice to have it.} {\color{blue} Added below.}

The computational complexity results are shown in Theorem~\ref{thm:thm2}.

\begin{theorem}\label{thm:thm2}
Consider the $t$th step of Algorithm~\ref{alg:MM-DUST} with $\lambda = \lambda_t$. The computational complexity is $\mathcal{O}(N_mp(n+N_dm))$ in logistic regression and  $\mathcal{O}(N_mp(nk+N_dm))$ in Cox regression with $k$ being the number of observed events.
\end{theorem}

We remark that the complexity is dominated by the number of majorization steps, $N_m$, while the complexity of the minimization step is always $\mathcal{O}(N_dmp)$ for all loss functions. With a fixed value of $\lambda$, gradient/sub-gradient based methods typically have computational complexity of $\mathcal{O}(np T)$, where $T$ is the total number of iterations~\citep{chen2012smoothing}. With large-scale datasets, our proposed MM-DUST algorithm can significantly reduce the computational time by using a small $N_m$ value. The efficiency can be resulted from the fact that the dual problem has reduced dimensions and the dual updates are not affected by the sample size $n$.

\subsection{Implementation Issues}

There are several key aspects of applying the MM-DUST algorithm, and we discuss them in the following sections. 

\subsubsection{Initial Value}

An initial estimate of $\bbeta$ is required to start the first majorization step, calculate the initial $\u$, and give the largest $\lambda$ value in the sequence. From the theoretial derivations and the numerical experiments, we have found that the algorithm is sensitive to the initial values. Since the path algorithm starts at the fully-penalized end of the primal problem, getting the initial $\bbeta$ amounts to solving the following optimization problem
\begin{align}
    \mathop{\mbox{minimize}}\limits_{\bbeta}\ f(\bbeta), \mbox{ s.t. } \|\D\bbeta\|_1=0. 
    \label{eq:MM-DUSTinital}
\end{align}

When $\D$ is of full column rank, we will have $\hat\bbeta_0=\bzero$. When $\D$ is not of full column rank, the optimal $\hat\bbeta_0$ may not be a zero vector. Instead, the solution should be in the \emph{Null} space of $\D$. Let $\V=(\v_1, \cdots, \v_q)\in\mathbb{R}^{p\times q}$ be a set of basis in the \emph{Null} space of $\D$ with $q$ being the nullity of $\D$. Problem~\eqref{eq:MM-DUSTinital} is then equivalent to 
\begin{align}
    \mathop{\mbox{minimize}}\limits_{\s}\ f(\V\s),
    \label{eq:MM-DUSTinital2}
\end{align}
where $\s\in\mathbb{R}^{q}$. In practice, the $\V$ matrix can be found by doing a singular value decomposition on $\D$. Besides, we will add a small $\ell_2$ penalty on $\s$ and take the solution from ridge regression when certain singular conditions happen in problem~\eqref{eq:MM-DUSTinital2}.

%\subsection{Standardization and Unstandardization}

\subsubsection{Degrees of Freedom and Early Stopping}

In real applications, the early stopping technique is widely applied with path algorithms to enhance computational efficiency. The information criteria, AIC, BIC, and GIC are commonly considered for model selection, which balance the model complexity and the goodness of fit. It is typical to use the degrees of freedom to describe the complexity of the model. 

Under the classical settings of linear regression with the normal error assumptions, the generalized lasso problem takes the form
\begin{align}
    \mathop{\mbox{minimize}}\limits_{\bbeta}\ \frac{1}{2}\|\y - \X\bbeta\|^2 + \lambda \|\D\bbeta\|_1. \label{eq:lsprob}
\end{align}
The degrees of freedom results are derived in \citet{tibshirani2011solution}, which is stated in the next proposition. 

\begin{proposition}\label{thm:df}
Suppose that $\rank(\X)=p$. For fixed $\lambda$, the fit $\X\hat\bbeta_{\lambda}$ of the generalized lasso \eqref{eq:lsprob} has degrees of freedom 
\begin{align*}
    \mbox{df}(\X\hat\bbeta_{\lambda}) = \mbox{E}[\mbox{nullity}(\D_{-\mathcal{B}})],
\end{align*}
where $\mathcal{B}$ is the index set of the dimensions that hit the boundary in a dual solution $\u$. The matrix $\D_{-\mathcal{B}}$ is defined by removing the corresponding rows in $\D$ according to the set $\mathcal{B}$.    
\end{proposition}

As we approximate the original generalized lasso problem by a quadratic majorizer for each specific $\lambda$ value, we take the degrees of freedom of the approximated problem as an approximation of the true complexity of the model. %In our algorithm, the boundary is just the $\lambda$ value in the dual problem. 
The application of Proposition~\ref{thm:df} is then straightforward. 

In practice, we calculate the information criterion when $\lambda$ is updated, and we record the value of the information criterion whenever there is a change in the degrees of freedom. The path algorithm will terminate when the criterion keeps increasing along a consecutively growing sequence of the degrees of freedom. In our numerical studies, the AIC rule is shown to provide better model selection and prediction results.

\section{Simulation Studies}
\label{sec:glasso-simu}

We compare the MM-DUST algorithm with the competing methods on the convergence of the solution paths and the model prediction performance. As the solution of the generalized lasso problem under the least square settings has been discussed elaborately in \citet{tibshirani2011solution}, in the simulation studies we mainly focus on the comparison of the algorithm under the logistic regression and Cox regression settings. The \emph{Smoothing Proximal Gradient} (SPG) algorithm \citep{chen2012smoothing} is selected as the competitor due to its flexibility in handling a wide range of convex loss functions. For both methods, we standardize the features for optimization and unstandardize the estimates for model outputs and prediction.

\subsection{Convergence of the Solution Paths}

\subsubsection{The Case of $\D = \I$}
\label{sec:converge-logit}

We show that solution paths provided by MM-DUST will approach the exact solution paths as the step size $\varepsilon$ decreases toward zero when $\D = \I$.

An example is given under the logistic regression settings. We simulate 400 observations with 10 covariates, in which we have each row of the design matrix $\X$ generated as $\x\sim N_{10}(\bzero,\I_{10})$. The binary response $y$ is generated as $y \sim \mbox{Bernoulli}(q)$, where  $q  = \exp(\eta)/\{1+\exp(\eta)\}$, and $\eta=-4 + \x^{\trans}\b$ with $\b=(-3, 3, -2, 2, -1, 1, 0.5, 0, 0, 0)^{\trans}$.

We set $\D = \I_{10}$ in the generalized lasso problem, which reduces to a lasso penalized logistic regression. For primal parameter $\bbeta$, we compare the solution paths from MM-DUST to the exact paths provided by \texttt{glmnet} \citep{friedman2010regularization}. It is worth noting that the SPG algorithm only provides the solution to an approximated problem. Thus, we do not use solution paths from SPG for comparison.

\begin{figure}[t]
     \centering
     \begin{subfigure}[b]{0.24\textwidth}
         \centering
         \includegraphics[width=\textwidth, height=\textwidth]{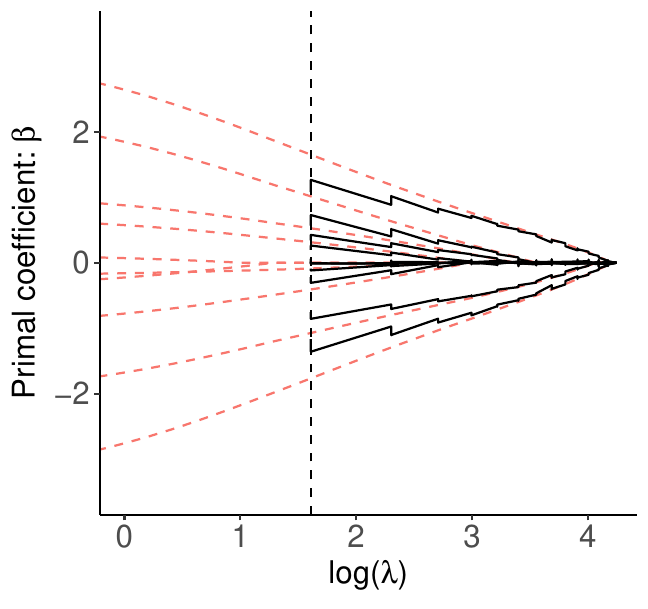}
         %\caption{$\varepsilon=5$}
     \end{subfigure}
     \hfill
     \begin{subfigure}[b]{0.24\textwidth}
         \centering
         \includegraphics[width=\textwidth, height=\textwidth]{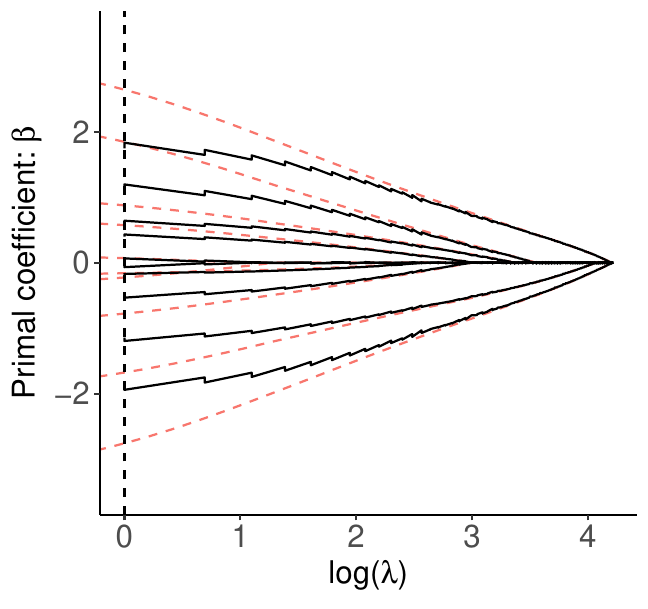}
         %\caption{$\varepsilon=1$}
      \end{subfigure}
     \begin{subfigure}[b]{0.24\textwidth}
         \centering
         \includegraphics[width=\textwidth, height=\textwidth]{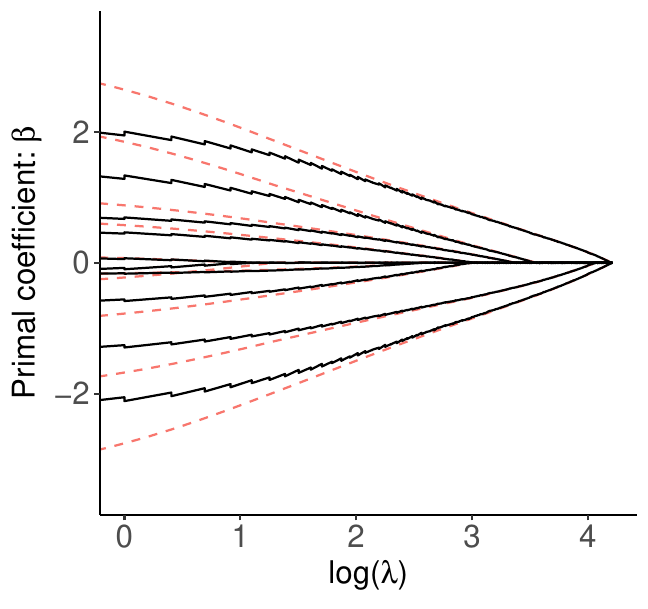}
         %\caption{$\varepsilon=0.5$}
      \end{subfigure}
      \hfill
     \begin{subfigure}[b]{0.24\textwidth}
         \centering
         \includegraphics[width=\textwidth, height=\textwidth]{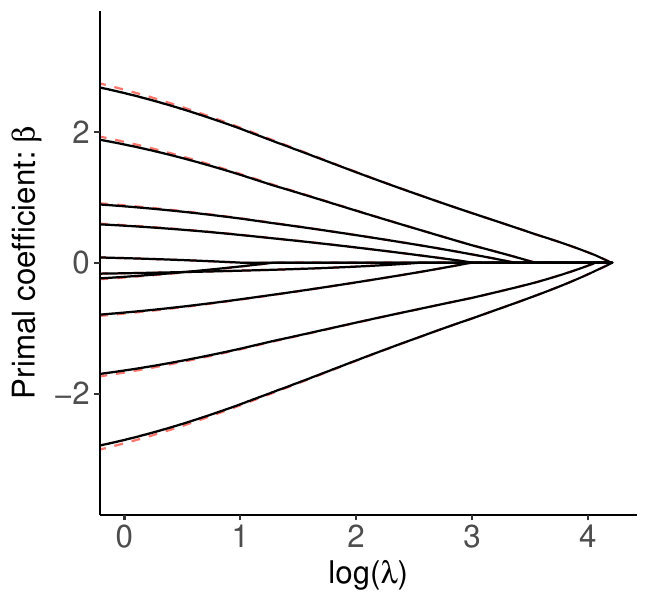}
         %\caption{$\varepsilon=0.01$}
      \end{subfigure}

      \begin{subfigure}[b]{0.24\textwidth}
         \centering
         \includegraphics[width=\textwidth, height=\textwidth]{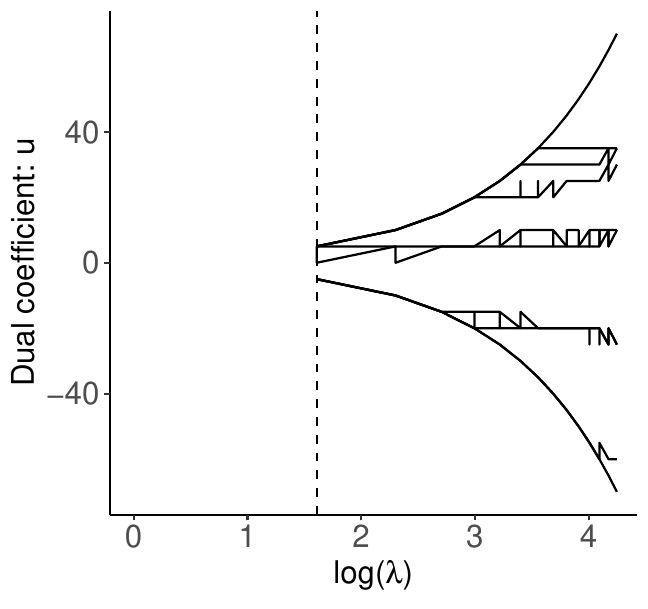}
         \caption{$\varepsilon=5$}
     \end{subfigure}
     \hfill
     \begin{subfigure}[b]{0.24\textwidth}
         \centering
         \includegraphics[width=\textwidth, height=\textwidth]{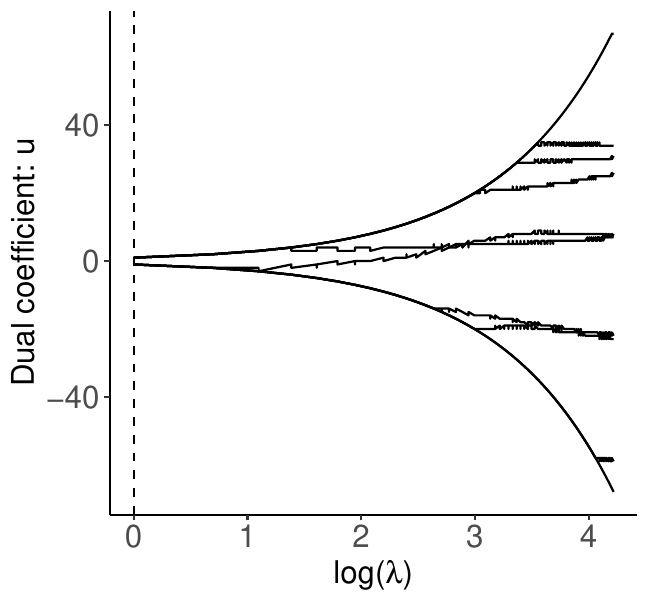}
         \caption{$\varepsilon=1$}
      \end{subfigure}
     \begin{subfigure}[b]{0.24\textwidth}
         \centering
         \includegraphics[width=\textwidth, height=\textwidth]{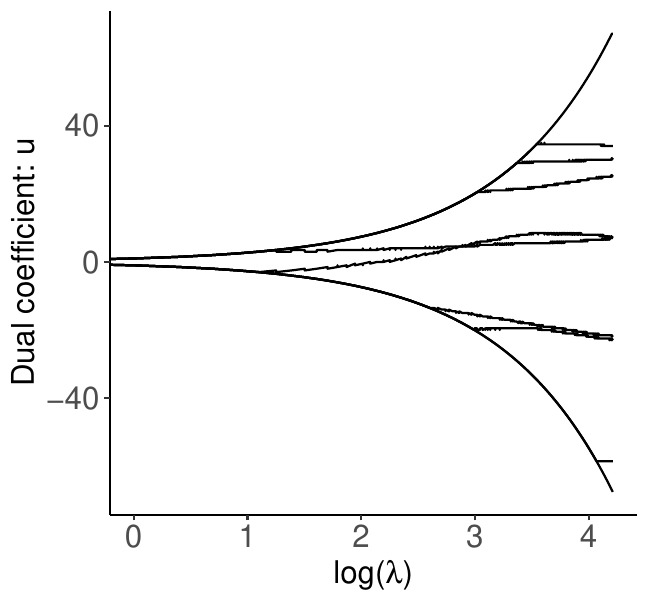}
         \caption{$\varepsilon=0.5$}
      \end{subfigure}
      \hfill
     \begin{subfigure}[b]{0.24\textwidth}
         \centering
         \includegraphics[width=\textwidth, height=\textwidth]{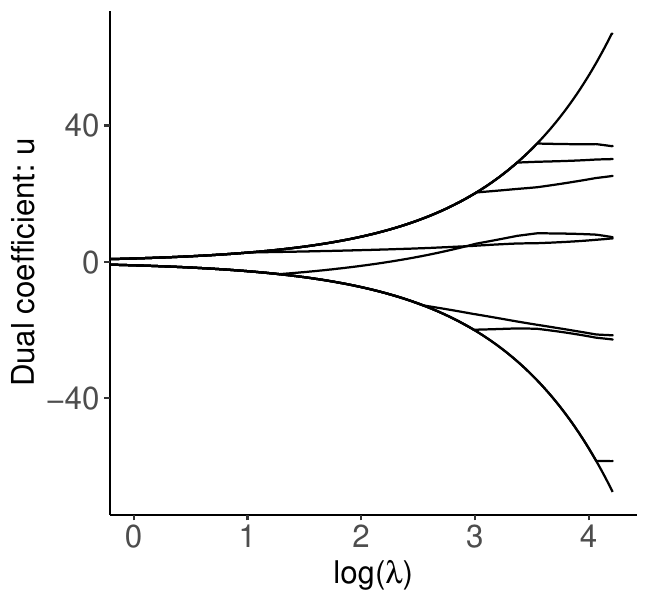}
         \caption{$\varepsilon=0.01$}
      \end{subfigure}

      \caption{Simulation: Solution paths of $\bbeta$ and $\u$ with varying step sizes. The exact solution paths from \texttt{glmnet} are shown in red dashed lines, while the paths from MM-DUST are shown in black solid lines. The top figures are paths for the primal coefficient $\bbeta$, while the bottom four are paths for the dual coefficient $\u$, with the x-axis as $\log(\lambda)$. The vertical dashed line marks the point when $\|\hat{\u}\|_{\infty}\leq \varepsilon$ and the algorithm stops. 
      %{\color{red} The labels are too small.
      %It may be better to zoom in to a smaller range, so the pattern can be clearer. How about $\lambda$ from 0 to 10?
      %}
      }
  \label{fig:path-logit}
\end{figure}

Figure~\ref{fig:path-logit} shows the zoomed-in solution paths for both the primal parameter $\bbeta$ and the dual parameter $\u$ with different step sizes $\varepsilon$. We set $N_m=5$ and $N_d=20$. The plots for the full solution paths are provided in Appendix~\ref{sec:app:fullpath}. 

The MM-DUST method approximately traces out the solution paths, and the disparities between MM-DUST and \texttt{glmnet} decrease as $\varepsilon$ becomes smaller. Interestingly, the step size $\varepsilon$ acts as a regularization parameter and larger $\varepsilon$ tends to shrink the coefficient more toward zero. With decreasing step sizes, both the primal paths and the dual paths grow smoother and the MM-DUST algorithm stops at a closer point toward the full model. Moreover, for any step size $\varepsilon$, the disparities between MM-DUST and \texttt{glmnet} are smaller for larger $\lambda$ values. In other words, MM-DUST tends to more accurately estimate sparser models; this is desirable for high-dimensional problems. 

%The last point in each MM-DUST path, which corresponds to the full model when $\lambda=0$, is given by the \texttt{glm} function in \emph{R}. 

\subsubsection{The Case of $\D\neq\I$}
\label{sec:converge-cox}

When $\D \neq \I$, we give an illustration under the Cox model. We simulate 400 observations with 10 covariates, where we have $\x\sim N_{10}(\bzero,\I_{10})$. The failure time $T$ is given by $T=-\log(u)/\{0.1\times \exp(\eta)\}$, where $u\sim Uniform(0, 1)$, $\eta = \x^{\trans}\b$, and $\b=(1, 1, 2, -2, -2, 3, 1.5, -0.5, 0, 0)^{\trans}$. The censoring time $C$ is sampled from an exponential distribution with rate parameter 0.9, and the final observed time $y$ is set as $\min(T, C)$. We take $\D\in\mathbb{R}^{13\times 10}$, in which the first three rows are set as 
\begin{align*}
\begin{bmatrix} 
1 & -1 & 0 & 0 & 0 & 0 & 0 & 0 & 0 & 0 \\
0 & 1 & -1 & 0 & 0 & 0 & 0 & 0 & 0 & 0 \\
0 & 0 & 0 & 1 & -1 & 0 & 0 & 0 & 0 & 0 \\
\end{bmatrix},
\end{align*}
and the rest ten rows at the bottom is $\I_{10}$. Note that, this $\D$ matrix is of full column rank. We set $N_m=5$ and $N_d=20$.

Under the settings of Cox regression with an arbitrary $\D$ matrix, there are few related algorithms that can directly be applied to get the exact solution paths. Thus, we report the solution paths from MM-DUST only. Based on $\D$, the regularization intends to shrink the differences $\beta_1-\beta_2$, $\beta_2-\beta_3$, and $\beta_4-\beta_5$ to 0. According to the true parameter $\b$, we expect the paths of $\beta_1$ and $\beta_2$, and the paths of $\beta_4$ and $\beta_5$ to be close. The path of $\beta_3$ can be close to the paths of $\beta_1$ and $\beta_2$ when $\lambda$ is large, but it will diverge when $\lambda$ decreases. Besides, the paths of $\beta_9$ and $\beta_{10}$ should be always close to 0. 

\begin{figure}[ht]
    \centering
    \includegraphics[width=\textwidth]{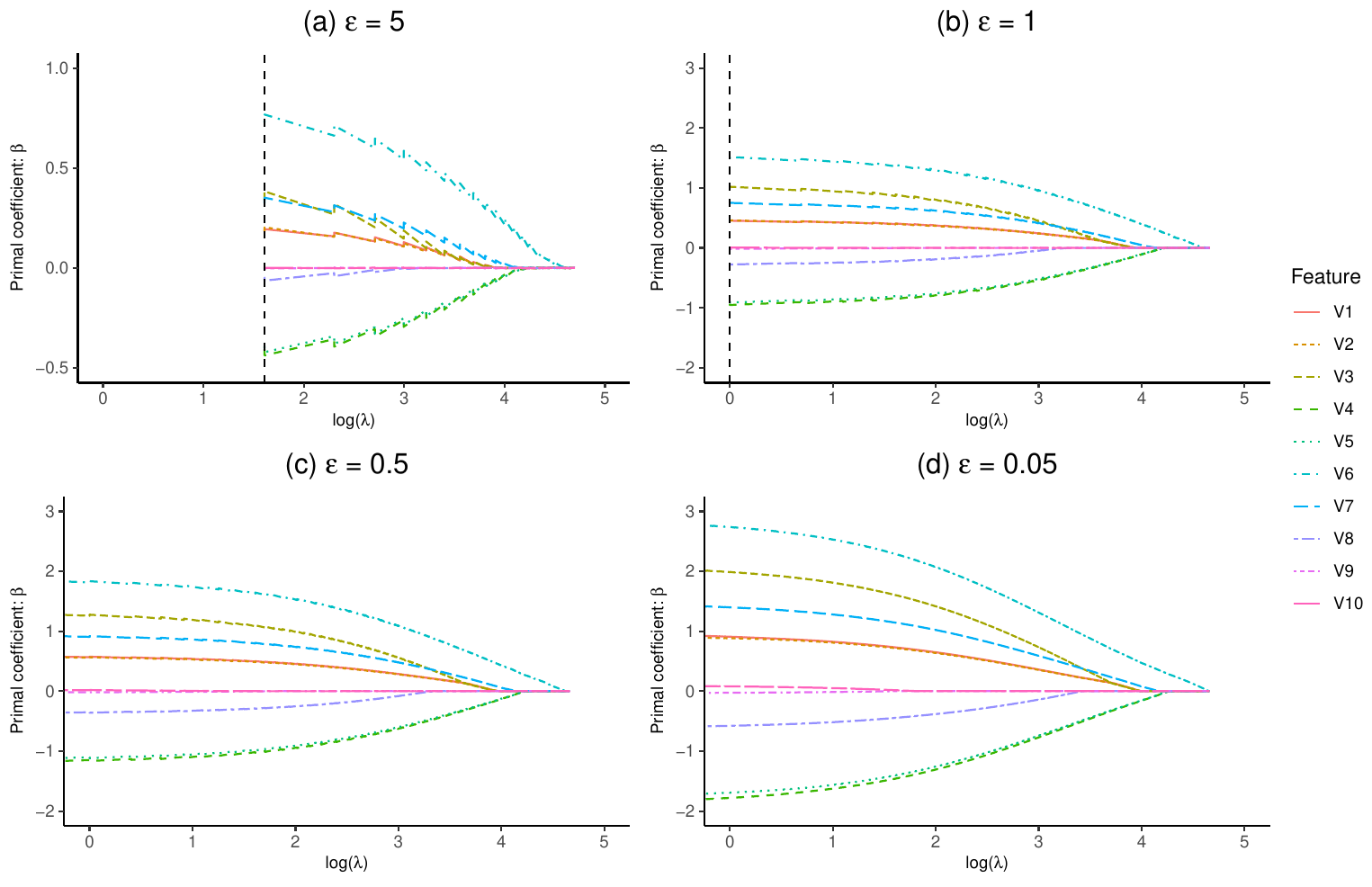}
    \caption{Simulation: Primal solution paths of $\bbeta$ for varying step sizes. The settings are the same as in Figure~\ref{fig:path-logit}. Note that (a) is further zoomed in on the y-axis to show the patterns clearer. %{\color{red} Same comment; make it clearer that the paths converges with smaller $\varepsilon$ and the sparsity is indeed obtained gradually; another idea is to show the graph with $\log(\lambda)$.}
    }
  \label{fig:path-cox}
\end{figure}

Figure~\ref{fig:path-cox} shows the zoomed-in primal paths with varied step sizes. Similar patterns are observed as in the previous example. The full paths are provided in Appendix~\ref{sec:app:fullpath}. 

\subsection{Model Performance Comparison}

We compare the prediction performance between the MM-DUST algorithm and the selected competing method, the SPG algorithm. For the MM-DUST algorithm, we use AIC for both model selection and early stopping. We consider four models from the MM-DUST algorithm: full path model with step size 0.1 (DUST(0.1)), full path model with step size 0.05 (DUST(0.05)), MM-DUST model of early stopping with step size 0.1 (DUST-A(0.1)), and MM-DUST model of early stopping with step size 0.05 (DUST-A(0.05)). In the early stopping models, the algorithm will terminate when the AIC keeps increasing over a sequence of 7 consecutive degrees of freedom values along the path. In all cases, models with the smallest AIC value will be selected. %We also include an elastic-net penalized Cox regression (Enet) as a benchmark. Since the Enet model solves a different sparse regression problem and the model is fitted by \texttt{glmnet}, the execution time is not comparable. 
Besides, we also report the results from the oracle Cox regression model with the true features (ORE). 

For the SPG method (SPG-cv), the best tuning parameter $\lambda$ is selected from 50 candidates via 5-fold cross validation. We adopt the warm start technique to speed up the calculation in SPG, and with each $\lambda$ value, the iteration will terminate when the absolute change in the objective function over the previous absolute objective function value is smaller than $10^{-4}$. %a tolerance parameter $\delta$.

We consider the generalized lasso problem based on the feature aggregation model proposed in \citet{yan2021rare}, which is applied when there is a given hierarchical structure among the predictors. A synthetic tree structure is provided in Figure~\ref{fig:simu_tree_cox} of Appendix~\ref{sec:app:tree}, and is used to construct the $\D$ matrix for the rest of the simulation studies. We remark that the resulted $\D$ matrix is always of full column rank. 

The simulations are conducted under the Cox regression settings, in which the evaluation of the gradient is computationally intensive. More details for the data generation process are left in Appendix~\ref{sec:app:simu}. 
%The constant $L$ required in the majorization step is calculated based on our derivations in Supplement~\ref{sec:coxproof}. 

For each of the simulation studies, we train the model with 300 observations and measure the model performance on 1000 independently sampled observations. Each simulation is repeated 100 times and we report the C-index values on the test sets and the execution times for model fitting.

\subsubsection{Effects of Signal-to-Noise Ratio}
\label{sec:glasso-simu-snr}

We investigate the performance of the MM-DUST algorithm under different signal-to-noise ratio (SNR): $\{0.5, 1, 1.5, 2\}$. In this study, the $\D$ matrix is of dimension $109\times 67$. That is, the dimension of the dual parameter $\u$ is $109$. This is the case when both the primal and the dual problem are low-dimensional. We take $N_m=1$ and $N_d=15$ in the simulations.
The prediction performance measures and the execution times are reported in Table~\ref{tab:mmcox}. 

%We consider four values of signal-to-noise ratio (SNR) $\in\{0.5, 1, 1.5, 2\}$ with $p=42$. By operating the tree-guided reparameterization, the final dimension of the primal parameter $\bgamma$ is $67$, and the $\D$ matrix is of dimension $109\times 67$. That is, the dimension of the dual parameter $\u$ is $109$. We take $N_m=1$ and $N_d=10$ in the simulations. The prediction performance measures and the execution times are reported in Table~\ref{tab:mmcox}. 

In all cases, the DUST(0.05) is the best MM-DUST model and exhibits comparable prediction performance, compared to the SPG-cv model, while significantly reducing the computational time. The early stopping models provide slightly lower C-index compared to models with the full paths traced out, but the differences get smaller when the step size gets smaller. The decrement in executing time is not substantial, which may be explained by the preference for a denser model over a sparser model under these model settings.

\begin{figure}[htp]
     \centering
     \begin{subfigure}[b]{0.45\textwidth}
         \centering
         \includegraphics[width=\textwidth, height=0.65\textwidth]{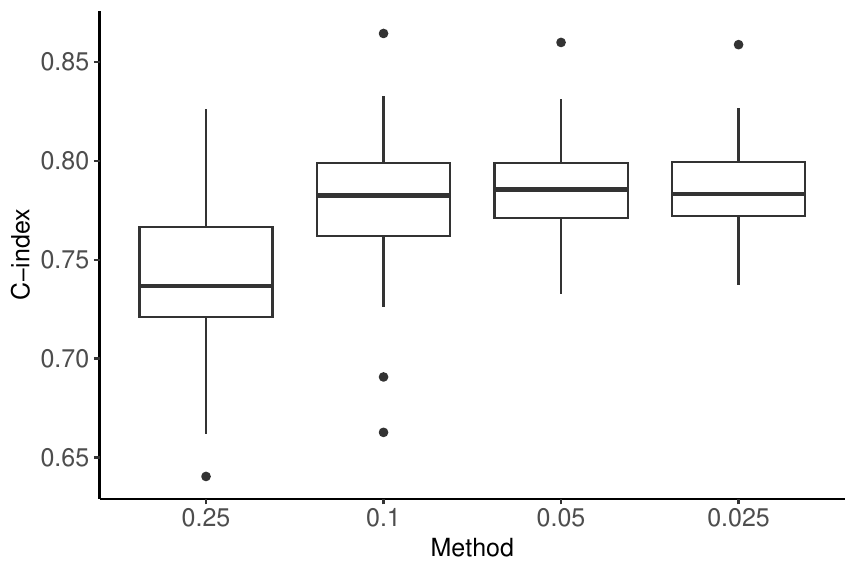}
         \caption{Prediction performance over step size}
     \end{subfigure}
     \begin{subfigure}[b]{0.45\textwidth}
         \centering
         \includegraphics[width=\textwidth, height=0.65\textwidth]{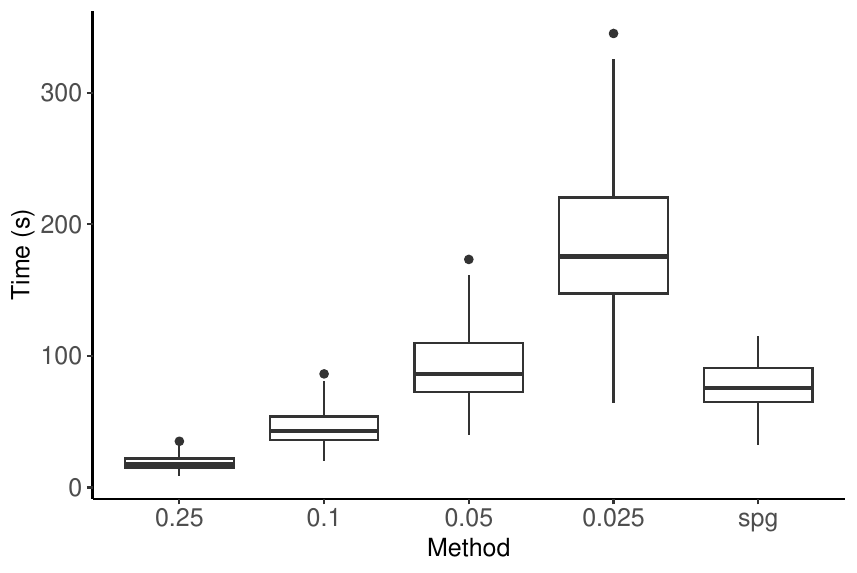}
         \caption{Computational time over step size}
      \end{subfigure}
      \caption{Simulation: comparison with varied step sizes and the SPG algorithm with SNR=1. Reported are the measures over 100 repetitions.}
  \label{fig:fullpath-cox}
\end{figure}

Moreover, we report the time to trace out the full solution paths by both SPG and MM-DUST when SNR = 1. For SPG, we find the solutions with the same $\lambda$ sequence as in MM-DUST when the step size $\varepsilon=0.25$. The metrics are reported in Figure~\ref{fig:fullpath-cox}. When step size reaches 0.05, the change in prediction accuracy is not substantial, but the increment in computational time is still non-negligible. Moreover, compared to MM-DUST, the SPG algorithm requires more time to generate the paths with an equal number of points.
%In all cases, the MM-DUST algorithm has shorter execution time than SPG.

Nevertheless, the simulations demonstrate that the MM-DUST algorithm can largely enhance computational efficiency while still giving comparable prediction performance by proper selection of the step size.

\begin{table}[h]
\caption{Simulation: prediction performance with different step sizes and SNR values. Reported are the means and standard errors over 100 repetitions.}
\label{tab:mmcox}
\centering
    \begin{subtable}[t]{0.46\textwidth}
        \centering
        \begin{tabular}[t]{lll}
            \toprule
             Method & C-index & Time (s)\\
              \cline{1-3}
              & \multicolumn{2}{c}{SNR = 0.5} \\
            \midrule
            ORE & 0.753 (0.001) & NA \\
            DUST(0.1) & 0.708 (0.003) & 47.4 (1.5)\\
            DUST(0.05) & 0.710 (0.002) & 94.8 (5.8)\\
            DUST-A(0.1) & 0.696 (0.003) & 42.2 (1.3)\\
            DUST-A(0.05) & 0.706 (0.002) & 86.2 (5.4)\\
            SPG-cv & 0.713 (0.002) & 207.9 (4.0)\\
            %Enet & 0.72 (0.002) & 8.174 (0.151) \\
            \bottomrule
        \end{tabular}
    \end{subtable}%
    \hspace{2mm}
    \begin{subtable}[t]{0.46\textwidth}
        \centering
        \begin{tabular}[t]{ll}
            \toprule
            C-index & Time (s)\\
            \cline{1-2}
            \multicolumn{2}{c}{SNR = 1} \\
            \midrule
            0.827 (0.001) & NA \\
            0.780 (0.003) & 44.4 (1.3)\\
            0.785 (0.002) & 89.2 (5.4)\\
            0.769 (0.003) & 40.5 (1.3)\\
            0.781 (0.002) & 82.4 (5.1)\\
            0.788 (0.002) & 197.8 (4.2)\\
            % 0.794 (0.002) & 9.65 (0.214)\\
            \bottomrule
        \end{tabular}
     \end{subtable}
     
    \begin{subtable}[t]{0.46\textwidth}
        \centering
        \begin{tabular}[t]{lll}
            \toprule
             & \multicolumn{2}{c}{SNR = 1.5} \\
            \midrule
            ORE & 0.867 (0.001) & NA \\
            DUST(0.1) & 0.823 (0.003) & 43.1 (1.3)\\
            DUST(0.05) & 0.827 (0.002) & 86.3 (5.2)\\
            DUST-A(0.1) & 0.810 (0.003) & 39.8 (1.2)\\
            DUST-A(0.05) & 0.821 (0.002) & 80.8 (5.1)\\
            SPG-cv & 0.830 (0.002) & 199.2 (4.5)\\
            %Enet & 0.835 (0.002) & 11.809 (0.33) \\
            \bottomrule
        \end{tabular}
    \end{subtable}%
    \hspace{2mm}
    \begin{subtable}[t]{0.46\textwidth}
        \centering
        \begin{tabular}[t]{ll}
            \toprule
            \multicolumn{2}{c}{SNR = 2} \\
            \midrule
            0.892 (0.001) & NA \\
            0.848 (0.003) & 42.0 (1.4)\\
            0.853 (0.002) & 83.6 (5.3)\\
            0.836 (0.003) & 39.7 (1.4)\\
            0.847 (0.002) & 78.7 (5.0)\\
            0.858 (0.002) & 204.3 (4.5)\\
            %0.861 (0.002) & 16.657 (0.966) \\
            \bottomrule
        \end{tabular}
     \end{subtable}
\end{table}

\subsubsection{Effects of Parameter Dimension}
\label{sec:glasso-simu-highd}

We also study the performance of MM-DUST under different dimensions of the parameters. For the matrix $\D$ of dimension $m\times p$, $p$ links to the dimension of the initial primal problem, while $m$ corresponds to the dimension of the dual problem. 

In the simulations, we test the MM-DUST algorithm with increased $p$ and $m$ with the feature aggregation model. Again, the dimension of $\D$ relies on the hierarchical tree structure among the features. Thus, we adjust the tree structure consecutively by the following steps to get $\D$ of different dimensions:  
\begin{itemize}
    \item $p=67$, $m=109$: Tree in Figure~\ref{fig:simu_tree_cox} of Appendix~\ref{sec:app:tree} with 42 leaf nodes.
    \item $p=131$, $m=231$: Add 5 nodes under the root node, each with 10 child nodes. Add another node with only 8 child nodes under the root node. Final tree is with 100 leaf nodes. 
    \item $p=186$, $m=336$: Add 5 nodes under the root node, each with 10 child nodes. Final tree is with 150 leaf nodes.
    \item $p=406$, $m=756$: Add 20 nodes under the root node, each with 10 child nodes. Final tree is with 350 leaf nodes.
\end{itemize}
For all the cases, the true data follows the same model with SNR = 1 as described in Section~\ref{sec:glasso-simu-snr}, but the case with a larger tree structure gets a more complicated regularization term. We still take $N_m=1$ and $N_d=15$. Besides, we report the results from SPG with a convergence tolerance of $10^{-3}$ in the last three cases, as it provides a much better prediction performance than the convergence tolerance of $10^{-4}$.

The model comparison results are presented in Table~\ref{tab:mmcox-high}. As the dimensions of $\D$ increase, the MM-DUST algorithm provides better prediction performance than the SPG algorithm, especially when $p$ or $m$ exceeds the sample size $n$. When the dimensions reach ($p=131,\ m=231$), the MM-DUST model selected with a step size of 0.1 already outperforms the model selected through cross validation from SPG, and the differences become larger as the dimensions increase. Moreover, the computational time of SPG grows rapidly with growing dimensions. The execution time of MM-DUST also goes up. However, as suggested by Table~\ref{tab:mmcox-high}, a significant amount of time can be saved by either using a larger step size, or implementing the early stopping technique, with slight sacrifice on the prediction performance. 
Additionally, the time reduction achieved through early stopping becomes more pronounced in high-dimensional scenarios. 

To summarize, the results demonstrate the strength of our algorithm in high-dimensional generalized lasso problem and suggest that potential benefits in prediction may also be obtained with proper step size selection.

\begin{table}[h]
\caption{Simulation: prediction performance with different step size and SNR. Reported are the means and standard errors over 100 repetitions. %{\color{red}I think it is better to only keep one or two digits in reporting time; this applies to all tables; you decide.}{\color{blue} Adjusted to 1 digits for all tables.}
} 
\label{tab:mmcox-high}
\centering
    \begin{subtable}[t]{0.46\textwidth}
        \centering
        \begin{tabular}[t]{lll}
            \toprule
             Method & C-index & Time (s)\\
              \cline{1-3}
              & \multicolumn{2}{c}{$(p=67,\ m=109)$} \\
            \midrule     
            ORE & 0.827 (0.001) & NA \\
            DUST(0.1)  & 0.780 (0.003) & 44.4 (1.3)\\
            DUST(0.05) & 0.785 (0.002) & 89.2 (5.4)\\
            DUST-A(0.1)  & 0.769 (0.003) & 40.5 (1.3)\\
            DUST-A(0.05) & 0.781 (0.002) & 82.4 (5.1)\\
            SPG-cv & 0.788 (0.002) & 197.8 (4.2)\\
            %Enet & 0.794 (0.002) & 9.65 (0.214)\\
            \bottomrule
        \end{tabular}
    \end{subtable}%
    \hspace{2mm}
    \begin{subtable}[t]{0.46\textwidth}
        \centering
        \begin{tabular}[t]{ll}
            \toprule
            C-index & Time (s)\\
            \cline{1-2}
            \multicolumn{2}{c}{$(p=131,\ m=231)$} \\
            \midrule
            0.830 (0.001) &  NA \\
            0.768 (0.003) & 155.8 (7.3)\\
            0.770 (0.002) & 254.7 (9.0)\\
            0.761 (0.003) & 85.1 (2.5)\\
            0.769 (0.002) & 171.8 (4.9)\\
            0.766 (0.003) & 285.4 (5.6)\\
            %0.766 (0.002) & 232.069 (5.303)\\
            \bottomrule
        \end{tabular}
     \end{subtable}
     
    \begin{subtable}[t]{0.46\textwidth}
        \centering
        \begin{tabular}[t]{lll}
            \toprule
             & \multicolumn{2}{c}{$(p=186,\ m=336)$} \\
            \midrule
            ORE & 0.829 (0.001) & NA \\
            DUST(0.1) & 0.761 (0.003) & 302.0 (8.0)\\
            DUST(0.05) & 0.761 (0.002) & 447.2 (11.4)\\
            DUST-A(0.1) & 0.754 (0.003) & 119.2 (4.1)\\
            DUST-A(0.05) & 0.757 (0.003) & 248.0 (8.2)\\
            SPG-cv & 0.756 (0.003) & 528.0 (11.1)\\
            %Enet & 0.748 (0.003) & 249.276 (4.771)\\
            \bottomrule
        \end{tabular}
    \end{subtable}%
    \hspace{4mm}
    \begin{subtable}[t]{0.46\textwidth}
        \centering
        \begin{tabular}[t]{ll}
            \toprule
            \multicolumn{2}{c}{$(p=406,\ m=756)$} \\
            \midrule
            0.829 (0.001) & NA \\
            0.740 (0.003) & 483.3 (15.8)\\
            0.740 (0.002) & 922.1 (25.3)\\
            0.729 (0.006) & 340.7 (10.7)\\
            0.741 (0.002) & 698.8 (20.7)\\
            0.719 (0.003) & 1135.5 (25.0)\\
            %0.724 (0.003) & 59.302 (1.12)\\
            \bottomrule
        \end{tabular}
     \end{subtable}
\end{table}

\section{Application}
\label{sec:glasso-real}

In this section, we investigate the performance of MM-DUST on two real datasets for the feature selection and aggregation problem~\citep{yan2021rare} under the classification settings. Unlike the simulation studies in which a multivariate normal distribution is assumed for $\x$, we have a design matrix of counts data in the first application and binary data in the second application. Specifically, in the second application, we explore the algorithm performance when the design matrix is extremely sparse with imbalanced response and a relatively large sample size. This scenario often results in substantial computational cost, underscoring the need for efficient algorithms to manage such challenging data structure.

\subsection{TripAdvisor Rating Prediction}

In the first application, we analyze the data contained in the \emph{R} package \texttt{rare}, originally sourced from \emph{TripAdvisor}. The predictor matrix, $\X$, is a 500-by-200 document-term matrix, where each row stands for a review and each column represents one of the 200 adjectives. The elements of $\X$ contain the frequency counts of each adjective within a given review. After removing 38 columns consisting entirely of zeros, the final predictor matrix is reduced to 500-by-162. The initial response is the hotel rating, which ranges from 1 to 5. We approach this as a binary classification problem, considering a high score as a rating of 4 or higher. This results in 81 reviews being categorized as high-score reviews.

A hierarchical clustering results of the 200 adjectives is provided in the package, which is obtained by first applying word embeddings on the adjectives and then conducting a hierarchical clustering of the word vectors. As the words with close meanings may have similar effects on the response, feature aggregation of the original features guided by the hierarchical structure is considered. We use the clustering result to generate the $\D$ matrix in the generalized lasso problem and the resulted $\D$ is a 521-by-359 matrix of full column rank. 

We evaluate the out-of-sample prediction performance of all methods with a 10-fold splitting procedure. The observations are randomly divided into 10 folds, and each time we use 9 folds for training and one fold for testing. For MM-DUST, we set $N_m=1$ and $N_d=20$, with the AIC models terminating when the AIC increases over 7 consecutive degrees of freedom. For SPG, a sequence of 50 $\lambda$ values are used with 5-fold cross validation. The out-of-sample AUCs and the execution times are shown in Table~\ref{tab:tripadvisor}. The DUST(0.1) model provides the highest AUC with the execution time reduced by almost 60\% and 84\% when compared to SPG-cv-3 and SPG-cv-4.

\begin{table}[H]
\caption{TripAdvisor rating prediction: reported are the means and standard errors of AUC over the 10 folds.  SPG-cv-3 and SPG-cv-4 stand for SPG model with a convergence tolerance of $10^{-3}$ and $10^{-4}$, respectively.}
\label{tab:tripadvisor}
\centering
\begin{tabular}[t]{l|ll}
\toprule
  & AUC & Time (s)\\
\midrule
DUST(0.1) & 0.643 (0.036) & 205.7 (16.8)\\
DUST(0.01) & 0.617 (0.022) & 2162.5 (117.8)\\
DUST-A(0.1) & 0.629 (0.037) & 162.1 (10.9)\\
DUST-A(0.01) & 0.621 (0.027) & 1656.6 (111.1)\\
SPG-cv-3 & 0.624 (0.019) & 521.9 (47.2)\\
SPG-cv-4 & 0.631 (0.030) & 1272.8 (114.3)\\
%GLM & 0.63 (0.025) & 0.226 (0.01)\\
%Enet & 0.669 (0.037) & 13.749 (0.987)\\
\bottomrule
\end{tabular}
\end{table}

\subsection{Suicide Risk Prediction}

In the second application, we use an electronic health records (EHR) dataset introduced in \citet{chen2024tree} for suicide risk prediction. %While the first application using counts data, features in the second application are binary and sparse, with a larger sample size and a more imbalanced response. 
In this study, the outcome of interest is the occurrence of suicide attempts during the study period. The dataset consists of 13,398 patients, in which 1,218 have committed suicide.

The features are derived from historical diagnosis recorded by the International Classification of Disease (ICD-10) codes under the ``F'' chapter, which is about mental, behavioral and neurodevelopmental disorders. Historical occurrence of the ICD-10 codes are aggregated and converted to binary predictors. We use a predictor set of 173 ICD-10 codes after a pre-screening process with an averaged prevalence of 0.54\%. Age and gender are also included as the predictors.

The ICD-10 codes are organized in a hierarchical structure, where the higher-level codes represent more generic disease categories and the lower-level codes represent more specific diseases and conditions. Consequently, we can consider feature aggregation guided by the hierarchical structure to deal with the severe sparsity issue of the design matrix. We construct the $\D$ matrix from the hierarchical structure of the ICD-10 codes and the resulted $\D$ matrix is 413-by-240.

%We use a predictor set of 173 ICD-10 codes after a pre-screening process with an averaged prevalence of 0.54\%. 
%ull-digit ICD-10 codes often results  Consequently, we can consider feature aggregation guided by the hierarchical structure to deal with the sparsity issue of the design matrix. 
%ICD-10 codes are organized in a hierarchical structure, where the higher-level codes represent more generic disease categories and the lower-level codes represent more specific diseases and conditions. 

The same 10-fold splitting procedure is employed. We set $N_m=5$ and $N_d=70$ with the AIC models terminating after 3 consecutive increases in AIC across degrees of freedom.  For SPG, we only use a sequence of 20 $\lambda$ values with 5-fold cross validation. The out-of-sample AUCs and the execution times are presented in Table~\ref{tab:suicideresultnew}. The MM-DUST models consistently achieve higher AUC values, though the differences are not substantial. On the other hand, although the dimensions of the $\D$ matrix are smaller than those of the first application, the increased sample size and sparse structure of $\X$ lead to a substantial rise in computational cost. With comparable prediction performance, DUST-A(10) model reduces computational costs by 90\% compared to SPG-cv-3, which has an averaged execution time of 38 hours. These findings indicate the advantages of MM-DUST for handling sparse features. Further, as suggested by our computational complexity analysis in Section~\ref{sec:comp}, the efficiency of MM-DUST with large-scale data is also enhanced as the dual updates are independent with the sample size. 

We remark that while reducing the convergence tolerance value or increasing the number of $\lambda$ values can boost prediction performance of SPG, the resulting substantial increase in the computational cost can overweight the benefits in prediction accuracy.

\begin{table}[H]
\caption{Suicide risk study: reported are the means and standard errors of AUC over the 10 folds.}
\label{tab:suicideresultnew}
\centering
\begin{tabular}[t]{l|ll}
\toprule
  & AUC & Time (m)\\
\midrule
DUST(10) & 0.722 (0.009) & 253.5 (17.4)\\
DUST(5) & 0.723 (0.009) & 312.0 (24.3)\\
DUST-A(10) & 0.722 (0.009) & 215.1 (16.8)\\
DUST-A(5) & 0.721 (0.010) & 286.7 (20.4)\\
SPG-cv-3 & 0.720 (0.009) & 2310.3 (180.6)\\
\bottomrule
\end{tabular}
\end{table}
\section{Discussion}
\label{sec:glasso-dis}

In this work, we have proposed a majorization-minimization dual stagewise (MM-DUST) algorithm for efficiently approximating the solution paths of the generalized lasso with a broad class of convex loss function. By formulating the model in the dual space and incrementally updating the dual coefficients with carefully chosen step sizes, the method provides a principled way to ``slow-brew'' the solution paths. We have established theoretical convergence guarantees and demonstrated, through simulation and real-world experiments, that MM-DUST achieves a compelling balance between computational efficiency and solution accuracy, especially for large-scale problems.
%The convergence of the proposed MM-DUST algorithm is established, and it offers a trade-off between computational efficiency and solution accuracy through choices of the step size and the number of dual iterations. Numerical studies demonstrate the efficacy of MM-DUST, especially with large-scale problems. %With appropriately selected step sizes, the algorithm closely mimics the exact paths and maintains satisfactory model fitting performance. %Given that the data sets in modern analysis are often very large-scale, it is important to explore more scalable computational algorithms or strategies for feature aggregation.  

There are several directions for future research. First, the overall computational time can be further reduced by incorporating \emph{adaptive step sizes}. Specifically, one could use a relatively larger step size at the beginning of the path to rapidly explore sparser models, and gradually shrink the step size as the model becomes denser. Second, although we focused on a stagewise learning framework, \emph{alternative optimization strategies} such as ADMM could be explored in the dual space if the ``slow-brewing'' feature is not suitable for specific applications. Last but not the least, it may be interesting to investigate extensions to other penalty structures, including overlapping group lasso or even non-convex penalties, as well as to explore the effects of more complex data-driven tuning strategies. 
%There are several directions for future research. Computational time can potentially be further reduced by incorporating an adaptive step size strategy. Intuitively, it is reasonable to use a relatively larger step size at the beginning of the paths to explore sparser models, while considering a smaller step size when the model becomes denser. On the other hand, instead of the stagewise algorithm, we can explore other algorithms such as ADMM to solve the dual problem when the ``slow-brewing'' property is not desired in some problems. Last but not the least, 

%\input{thm}

%\newpage
\section*{Supplementary Materials}

The online supplementary materials include an R implementation of the proposed approaches, derivations of computational algorithms, theoretical analysis, and supporting numerical results. 

%\newpage

\bibliographystyle{chicago}
\bibliography{reference}

\clearpage
\appendix
\noindent {\bf \LARGE Supplement}
\section{Derive the Dual Problem}\label{supp:dual}

Consider the generalized lasso problem: $\min_{\bbeta}\ f(\bbeta) + \lambda\|\D\bbeta\|_1$. 
We first augment the original problem by introducing a new vector $\z\in\mathbb{R}^{m}$ as the following
\begin{align*}
    \mathop{\mbox{minimize}}\limits_{\bbeta, \z}\ f(\bbeta) + \lambda \|\z\|_1, \mbox{ s.t. } \z = \D\bbeta.
\end{align*}
The Lagrangian form is then 
\begin{align}
    \mathop{\mbox{minimize}}\limits_{\bbeta, \z, \u}\ f(\bbeta) + \lambda \|\z\|_1 + \u^{\trans}(\D\bbeta-\z),
    \label{eq:primal}
\end{align}
where $\u\in\mathbb{R}^m$ is the Lagrangian multiplier. 

By definition of the dual problem, we have the dual as 
\begin{align}
    \mathop{\mbox{maximize}}\limits_{\u}\{\mathop{\mbox{minimize}}\limits_{\bbeta,\z} f(\bbeta)+ \lambda \|\z\|_1 + \u^{\trans}(\D\bbeta-\z)\}.
\end{align}
As the inner minimization problem is separable for $\bbeta$ and $\z$, the inner minimization is easy to be solved. The optimal $\bbeta$ is the solution to 
\begin{align*}
    \mathop{\mbox{minimize}}\limits_{\bbeta} f(\bbeta) + \u^{\trans}\D\bbeta = -\mathop{\mbox{maximize}}\limits_{\bbeta}\{-f(\bbeta) - \u^{\trans}\D\bbeta\} = -f^*(-\D^{\trans}\u),
\end{align*}
where $f^*(\cdot)$ is the convex conjugate of $f(\cdot)$ defined as $f^*(\y)=\max_{\x}\{\y^{\trans}\x-f(\x)\}$. On the other hand, for $\z$, we try to minimize $L(\z)=\lambda\|\z\|_1-\u^{\trans}\z$, and the optimal point satisfies $\bzero \in \partial L(\z)=\lambda\partial\|\z\| - \u$, where $\partial$ is the sub-gradient operator. Consequently, $\z=\bzero$ when $\|\u\|_{\infty}\leq \lambda$, and the minimum of $L(\z)$ is 0. Otherwise, $L(\z)=-\infty$ when $\|\u\|_{\infty}>\lambda$. To have a proper definition for $\min L(\z)$, we must have $\|\u\|_{\infty}\leq \lambda$.

Thus, the dual problem can be written as
\begin{align*}
    &\mathop{\mbox{maximize}}\limits_{\u}-f^*(-\D^{\trans}\u), \,\mbox{s.t. }\|\u\|_{\infty}\leq \lambda,\\
    \rightarrow
    &\mathop{\mbox{minimize}}\limits_{\u}f^*(-\D^{\trans}\u), \,\mbox{s.t. }\|\u\|_{\infty}\leq \lambda.
\end{align*}

\section{Proof of Lemma~\ref{lemma:dual}}
\label{app:glasso-proof}

In this section, we want to show that for $0<\lambda<\lambda_0$, a solution $\hat\u$ that satisfies the stopping rule in Algorithm~\ref{alg:dual} has suboptimality, that is, the objective function value $f(\hat\u)$ converges to the global minimum.

Without loss of generality, we write the box-constrained problem ~\eqref{eq:dual} as 
\begin{align}\label{eq:dualback}
    \mathop{\mbox{minimize}}\limits_{\u}\ f(\u), \mbox{ s.t. } \|\u\|_{\infty} \leq \lambda,
\end{align}
where $f$ is a convex, L-smooth, and differentiable function, and $\u\in\mathbb{R}^m$. 

Based on the results in \citet{jaggi2013revisiting}, for problem ~\eqref{eq:dualback}, a valid duality gap is
\begin{align*}
    h(\u) = \mathop{\mbox{maximize}}\limits_{\|\z\|_{\infty}\leq\lambda} \nabla f(\u)^{\trans}(\u-\z).
\end{align*}
When $\u^*$ is a solution of the problem, $f(\u)-f(\u^*)\leq h(\u)$ for all feasible points $\u$ such that $\|\u\|_{\infty}\leq\lambda$. Moreover, we can further simplify $h(\u)$ as
\begin{align*}
    h(\u) = \nabla f(\u)^{\trans}\u + \lambda \mathop{\mbox{maximize}}\limits_{\|\z\|_{\infty}\leq1}\{ \nabla f(\u)^{\trans}\z\} = \nabla f(\u)^{\trans}\u + \lambda\|f(\u)\|_1,
\end{align*}
where we use the property that $\|a\z\|_{\infty}=|a|\|\z\|_{\infty}$, $\forall a$, in the first equality, and use the fact that the dual norm of the infinity norm is the $\ell_1$ norm in the second equality. 

Suppose $\hat\u$ is the solution of Algorithm~\ref{alg:dual} when the stopping rule is satisfied. We define the index set $\mathcal{K}=\{j: |\hat\u_j| = \lambda, \, j=1, \ldots, m \}$. For all $j\in\mathcal{K}^c$, we have no descent direction with a step size of $\varepsilon$. It indicates that
\begin{align*}
    f(\hat \u \pm \varepsilon\bone_j) \geq f(\hat\u).
\end{align*}
Since $f$ is L-smooth, we have
\begin{align*}
    f(\hat\u\pm\varepsilon\bone_j) \leq f(\hat\u) \pm \varepsilon\nabla f(\hat\u)^{\trans}\bone_j + \frac{L}{2}\varepsilon^2.
\end{align*}
Combine the above two inequalities, we can conclude that
\begin{align}\label{eq:Kc}
    \mp \varepsilon\nabla f(\hat\u)^{\trans}\bone_j \leq \frac{L}{2}\varepsilon^2 \Rightarrow |\nabla_j f(\hat\u)| \leq \frac{L}{2}\varepsilon, \ \forall j\in\mathcal{K}^c.
\end{align}
On the other hand, for $j\in\mathcal{K}$, update in the feasible direction will increase the loss function. For the infeasible direction, a trivial case is that the update also increases the loss. Under such $j$, we still have inequality~\eqref{eq:Kc} satisfied. We focus on the case when the infeasible direction is a descent direction. %e descent direction is not feasible, that is, update in the descent direction with an $\varepsilon$ step will cross the boundary. 
Assume $\hat u_j=\lambda$, we have
\begin{align*}
    f(\hat\u + \varepsilon\bone_j) \leq f(\hat\u), \ 
    f(\hat\u + \varepsilon\bone_j) \geq f(\hat\u) + \varepsilon\nabla f(\hat\u)^{\trans}\bone_j,
\end{align*}
where the second inequality uses the first order condition of convexity. Therefore, we get $\nabla_j f(\hat\u)\leq 0$ for $\hat u_j=\lambda$. Similarly, when $\hat u_j=-\lambda$, we have $\nabla_j f(\hat\u)\geq 0$. It then follows that
\begin{align}\label{eq:K}
    \nabla_j f(\hat\u)\hat u_j + \lambda|\nabla_j f(\hat\u)| = 
    \begin{cases}
        \lambda \nabla_j f(\hat\u) + \lambda (-\nabla_j f(\hat\u)) = 0, \ &\hat u_j = \lambda; \\
        - \lambda \nabla_j f(\hat\u) + \lambda \nabla_j f(\hat\u) = 0, \ &\hat u_j = -\lambda.
    \end{cases}
\end{align}

Combine result~\eqref{eq:Kc} and result~\eqref{eq:K}, the duality gap at $\hat\u$ is
\begin{align*}
    h(\hat\u) 
    &= \nabla f(\hat\u_{\mathcal{K}_0^c})^{\trans}\hat\u_{\mathcal{K}_0^c} + \lambda\|\nabla f(\hat\u_{\mathcal{K}_0^c})\|_1 + \sum_{j\in\mathcal{K}_0}\nabla_j f(\hat\u)\hat u_j + \lambda|\nabla_j f(\hat\u)| \\
    &\leq \|\nabla f(\hat\u_{\mathcal{K}_0^c})^{\trans}\|_1\|\hat\u_{\mathcal{K}_0^c}\|_{\infty} + \lambda\|\nabla f(\hat\u_{\mathcal{K}_0^c})\|_1 + 0 \\
    &\leq 2\lambda |\mathcal{K}_0^c|\frac{L}{2}\varepsilon
    \leq L\lambda m\varepsilon,
\end{align*}
where $\mathcal{K}_0^c=\mathcal{K}^c\cup\{j\in \mathcal{K}: \eqref{eq:Kc}\mbox{ is satisfied}\}$.% is the number of dimensions in $\hat\u$ that is not on the boundary. 

Therefore, as $\varepsilon \rightarrow 0$, we have
\begin{align*}
    f(\hat\u) - f(\u^*) \leq h(\hat\u) \rightarrow 0.
\end{align*}
Since $f(\hat\u) - f(\u^*)\geq 0$ as $f(\u^*)$ is the global minimum, we can conclude that $f(\hat\u) \rightarrow f(\u^*)$ as $\varepsilon \rightarrow 0$.

Specifically, when $f(\hat\u) = \|y-\D^{\trans}\hat\u\|^2$, we have 
\begin{align*}
    \|y-\D^{\trans}\hat\u\|^2 - \|y-\D^{\trans}\u^*\|^2 = \|\D^{\trans}\hat\u-\D^{\trans}\u^*\|^2 + 2(y-\D^{\trans}\u^*)^{\trans}(\D^{\trans}\u^*-\D^{\trans}\hat\u).
\end{align*}
By definition of global minimum, we have
\begin{align*}
    2(y-\D^{\trans}\u^*)^{\trans}(\D^{\trans}\u^*-\D^{\trans}\hat\u) = \nabla f(\u^*)^{\trans}(\hat\u - \u^*) \geq 0.
\end{align*}
We can conclude that $\|\D^{\trans}\hat\u-\D^{\trans}\u^*\|^2 \rightarrow 0$ as $\varepsilon \rightarrow 0$. We remark that this result does not require strongly convexity for $f$, and the $\D$ matrix does not need to have full row rank.

\section{Proof of Lipschitz Constant in Cox Regression}
\label{sec:coxproof}

Let $\mathcal{D}$ be the index set of subjects that have the observed event, and $\mathcal{R}_s$ be the risk set of the $s$th event in the dataset. We consider the negative log partial likelihood when there are no ties as the loss function, where
\begin{align*}
            \ell(\bbeta) = \sum_{s\in\mathcal{D}}(-\x_s^{\trans}\bbeta + \log\sum_{i\in\mathcal{R}_s}\exp(\x_i^{\trans}\bbeta)).
\end{align*}
The goal is to find a diagonal matrix $\M$, such that $\M\succeq \nabla^2\ell(\bbeta),\ \forall \bbeta$, which is also simple in computation. 

%The next step amounts to finding such matrix $\M$ which is also simple in computation. 
%Denote the partial derivative of the negative log partial likelihood function
%$-\ell(\bnu)$ with respect to $\bnu$ as $g(\bnu)$. 
First of all, we have
\begin{align*}
  %g(\bnu) = 
  \ell^{'}(\bbeta) = %\frac{\partial \ell(\bnu)}{\partial \bnu}
    \sum_{s\in \mathcal{D}} \Big( - \x_s + \frac{\sum_{i\in \mathcal{R}_s} \x
_i \exp(\x_i^{\trans}
    \bbeta)}
    {\sum_{i\in \mathcal{R}_s} \exp(\x_i^{\trans} \bbeta)} \Big). \\
\end{align*}
%We further define
%\begin{align*}
%  &\M = \diag \Big(\sum_{s\in \mathcal{D}}\frac{1}{4}(\max_{i\in R_s} w_{ij} -
%  \min_{i\in R_s} w_{ij})^2\Big)_{j\in \{1, \ldots, |T|+q\}} .
%\end{align*}
Define $\ell_s(\bbeta) = -\x_s^{\trans}\bbeta + \log \sum_{i\in \mathcal{R}_s} \exp(\x_i^{\trans} \bbeta)$, then $\ell(\bbeta) = \sum_{s\in \mathcal{D}}\ell_s(\bbeta)$. We have the $j$th element in the first derivative of $\ell_s(\bbeta)$ as 
\begin{align*}
     \ell_{sj}^{'}(\bbeta) 
    & = -x_{sj}+\frac{\sum_{i\in \mathcal{R}_s}x_{ij}exp(\x_i^{\trans}\bbeta)}{\sum_{i\in \mathcal{R}_s}exp(\x_i^{\trans}\bbeta)}. \\
\end{align*}
Similarly, we can also conduct the decomposition for $\nabla^2\ell(\bbeta)=\sum_{i\in \mathcal{R}_s}\nabla_s^2\ell(\bbeta)$. 
The element at the position $(j, k)$ in the matrix $\nabla_s^2\ell(\bbeta)$ can be expressed as 
\begin{align*}
    \ell_{sjk}^{''}(\bbeta)
    & = \frac{\sum_{i\in \mathcal{R}_s}x_{ij}x_{ik}exp(\x_i^{\trans}\bbeta)\times \sum_{i\in \mathcal{R}_s}exp(\x_i^{\trans}\bbeta) }{(\sum_{i\in \mathcal{R}_s}exp(\x_i^{\trans}\bbeta))^2} - 
    \frac{\sum_{i\in \mathcal{R}_s}x_{ij}exp(\x_i^{\trans}\bbeta)}{\sum_{i\in \mathcal{R}_s}exp(\x_i^{\trans}\bbeta)}\times 
    \frac{\sum_{i\in \mathcal{R}_s}x_{ik}exp(\x_i^{\trans}\bbeta)}{\sum_{i\in \mathcal{R}_s}exp(\x_i^{\trans}\bbeta)} \\
    & = \sum_{i\in \mathcal{R}_s}\left(x_{ij}-\frac{\sum_{i\in \mathcal{R}_s}x_{ij}exp(\x_i^{\trans}\bbeta)}{\sum_{i\in \mathcal{R}_s}exp(\x_i^{\trans}\bbeta)}\right)
    \left(x_{ik}-\frac{\sum_{i\in \mathcal{R}_s}x_{ik}exp(\x_i^{\trans}\bbeta)}{\sum_{i\in \mathcal{R}_s}exp(\x_i^{\trans}\bbeta)}\right)
    \frac{exp(\x_i^{\trans}\bbeta)}{\sum_{i\in \mathcal{R}_s}exp(\x_i^{\trans}\bbeta)} \\
    & \triangleq \sum_{i\in \mathcal{R}_s}a_{ij}a_{ik}p_i,\\
    \ell_{sjj}^{''}(\bbeta) 
    & = \frac{\sum_{i\in \mathcal{R}_s}x_{ij}^2exp(\x_i^{\trans}\bbeta)}{\sum_{i\in \mathcal{R}_s}exp(\x_i^{\trans}\bbeta)}-
    \left(\frac{\sum_{i\in \mathcal{R}_s}x_{ij}exp(\x_i^{\trans}\bbeta)}{\sum_{i\in \mathcal{R}_s}exp(\x_i^{\trans}\bbeta)}\right)^2.
\end{align*}

Consider $\M_s=\diag(m_{s1},m_{s2},\ldots,m_{sp})$ and $\M=\sum_{s\in\mathcal{D}}\M_s$. Then for $\forall \z\in\mathbb{R}^p$, 
\begin{align*}
    \z^{\trans}(\M-(-\nabla^2\ell(\bbeta))\z
    & = \sum_{s\in \mathcal{D}}(\z^{\trans}\big(\M_s-(-\nabla_s^2\ell(\bbeta))\z\big) \\
    & = \sum_{s\in \mathcal{D}}\big(\sum_{j=1}^p(m_{sj}-\ell_{sjj}^{''}(\bbeta))z_j^2 + 
    \sum_{j=1}^pz_j\sum_{k\neq j}^pz_k(-\sum_{i\in \mathcal{R}_s}a_{ij}a_{ik}p_i)\big)\\
    & = \sum_{s\in \mathcal{D}}\big(\sum_{j=1}^p(m_{sj}-\ell_{sjj}^{''}(\bbeta))z_j^2 -
    \sum_{i\in \mathcal{R}_s}p_i
    \sum_{j=1}^pz_ja_{ij}\sum_{k\neq j}^pz_ka_{ik}\big) \\
    & = \sum_{s\in \mathcal{D}}\big(\sum_{j=1}^p(m_{sj}-\ell_{sjj}^{''}(\bbeta))z_j^2 - 
    \sum_{i\in \mathcal{R}_s}p_i
    (\sum_{j=1}^pz_ja_{ij})^2 + 
    \sum_{i\in \mathcal{R}_s}p_i
    \sum_{j=1}^pz_j^2a_{ij}^2 \big) \\
    & = \sum_{s\in \mathcal{D}}\big(\sum_{j=1}^p(m_{sj}-\ell_{sjj}^{''}(\bbeta) + \sum_{i\in \mathcal{R}_s}p_ia_{ij}^2)z_j^2 - 
    \sum_{i\in \mathcal{R}_s}p_i
    (\sum_{j=1}^pz_ja_{ij})^2 \big) \\
    & \geq \sum_{s\in \mathcal{D}}\big(\sum_{j=1}^p(m_{sj}-\ell_{sjj}^{''}(\bbeta) + \sum_{i\in \mathcal{R}_s}p_ia_{ij}^2)z_j^2 - 
    \sum_{i\in \mathcal{R}_s}p_i
    \sum_{j=1}^pz_j^2\sum_{j=1}^pa_{ij}^2 \big) \\
    & = \sum_{s\in \mathcal{D}}\big(\sum_{j=1}^p(m_{sj}-\ell_{sjj}^{''}(\bbeta) + \sum_{i\in \mathcal{R}_s}p_ia_{ij}^2- \sum_{i\in \mathcal{R}_s}p_i\sum_{j=1}^pa_{ij}^2)z_j^2 \big).
\end{align*}
The inequality follows from the Cauchy-Schwarz inequality.
It is not hard to show that $\ell_{sjj}^{''}(\bbeta)=\sum_{i\in R_s}p_ia_{ij}^2$. Finally, we get
\begin{align*}
    \z^{\trans}(\M-\nabla^2\ell(\bbeta))\z 
     & \geq 
     \sum_{s\in \mathcal{D}} \big( \sum_{j=1}^p(m_{sj}- \sum_{i\in \mathcal{R}_s}p_i\sum_{j=1}^pa_{ij}^2)z_j^2 \big) 
    = \sum_{s\in \mathcal{D}} \big(\sum_{j=1}^p(m_{sj}- \sum_{j=1}^p\ell_{sjj}^{''}(\bbeta))z_j^2 \big) \\
    & = \sum_{j=1}^p \big(\sum_{s\in \mathcal{D}}(m_{sj}- \sum_{j=1}^p\ell_{sjj}^{''}(\bbeta))z_j^2 \big) 
    = \sum_{j=1}^p \big(m_j- \sum_{j=1}^p\ell_{jj}^{''}(\bbeta) \big) z_j^2, 
\end{align*}
where $m_j$ is the $j$th diagonal element in $\M$. 

As it is proved in \citet{yang2012cmd},  
\begin{align*}
  m_j^* =  \sum_{s\in \mathcal{D}}\frac{1}{4}(\max_{i\in \mathcal{R}_s} x_{ij} -
  \min_{i\in \mathcal{R}_s} x_{ij})^2 \geq \ell_{jj}^{''}(\bbeta).
\end{align*}
Therefore, we can take $m_j=\sum_{j=1}^p m_j^*$ and the resulted $\M$ matrix will always satisfy the condition. In practice, we may also use $m_j=c\max{m_j^*}$ where $c$ is a user-defined constant.

\section{Convergence and Complexity Results for MM-DUST}\label{app:sec:conv}

\subsection{Complexity}

Assume that design matrix is $n$-by-$p$, and we initialize the $t$th iteration at $\hat\bbeta_{t-1}$ with $\lambda = \lambda_t$. We mainly conduct three steps: the majorization step to update surrogate function, the minimization step via Dual-Solver, and the primal-dual updating step. 

\begin{itemize}
    \item Majorization step: For the majorization step, we need to calculate the gradient of the original primal problem $\nabla f(\hat\bbeta_{t-1})$ and the current objective function value. Under logistic regression, the evaluation of $\X\hat\bbeta_{t-1}$ dominates the computational time, and the computational complexity is $\mathcal{O}(np)$. For the Cox regression,  we assume there are $k$ events in the data. Additional to the evaluation of $\X\hat\bbeta_{t-1}$, the gradient and the loss function involve the risk set searching for each event. Thus, the complexity in the worst case is $\mathcal{O}(npk)$.
    \item Minimization step: For the minimization step, the computational time can be significantly saved as the dual updates do not involve sample size $n$. The design matrix of the dual least-square problem is of dimension $p$-by-$m$. Thus, the complexity to get the change with $\varepsilon$ step update in each direction is $\mathcal{O}(mp)$, and the complexity to find the largest decrement is $\mathcal{O}(m)$. As we take $N_d$ dual steps, the final complexity is $\mathcal{O}(N_dmp)$.
    \item Primal-dual updating step: After the minimization step, we use formula~\eqref{eq:primal-dual2} to update the primal coefficient, in which the dominate step is to evaluate $\D^{\trans}\hat\u$. The complexity is therefore $\mathcal{O}(mp)$.
\end{itemize}

Each majorization step will be followed by a minimization step and a primal-dual update. Overall, when we take $N_m$ majorization step in the $t$th iteration, the computational complexity is $\mathcal{O}(N_m(np + N_dmp))$ under logistic regression and $\mathcal{O}(N_m(npk + N_dmp))$ under Cox regression.

\subsection{Lemma~\ref{lemma2} and Proof}

Before giving the convergence results for the MM-DUST algorithm, we first prove the following Lemma. 
\begin{lemma}\label{lemma2}
The initial primal parameter $\hat\bbeta_0$ defined in Algorithm~\ref{alg:MM-DUST} is the solution to 
\begin{align*}
    \mathop{\mbox{minimize}}\limits_{\bbeta} f(\bbeta) + \lambda_0\|\D\bbeta\|_1,
\end{align*}
where $f(\cdot)$ is strictly convex and $\lambda_0=\|\hat\u_0\|$ as defined in Algorithm~\ref{alg:MM-DUST} before rounding.
\end{lemma}

Based on the design of Algorithm~\ref{alg:MM-DUST}, at the initialization step, we have the initial primal parameter as
\begin{align}\label{eq:iniprob1}
    \hat\bbeta_0  = \arg\min f(\bbeta), \ \mbox{s.t. } \D\bbeta = 0,
\end{align}
which can be explicitly solved when $f(\cdot)$ is strictly convex. 
Based on $\hat\bbeta_0$, we further construct a surrogate primal objective function as the following
\begin{align}\label{eq:iniprob2}
    g_0(\bbeta) = \frac{1}{2L}\|L\hat\bbeta_0 - \nabla f(\hat\bbeta_0) - L\bbeta\|^2, \ \mbox{s.t. } \D\bbeta = 0.
\end{align}
We denote the solution that minimizes problem~\eqref{eq:iniprob2} as $\tilde\bbeta_0$.

First of all, we show that $\hat\bbeta_0 = \tilde\bbeta_0$. Let $\V=(\v_1, \cdots, \v_q)$ as a $p\times q$ matrix with columns being a set of orthogonal basis for the \emph{Null} space of $\D$. Then solving problem~\eqref{eq:iniprob1} is equivalent to find a $q$-dimensional vector $\hat\s_0$, such that $\hat\bbeta_0=\V\hat\s_0$ and $\hat\s_0 = \arg\min f(\V\s)$. From the optimality condition, we have  
\begin{align}
    \nabla f(\hat\s_0) = \V^{\trans}\nabla f(\hat\bbeta_0) = \bzero.
\end{align}
Similarly, solving problem~\eqref{eq:iniprob2} is equivalent to find a $\tilde\s_0\in\mathbb{R}^q$, such that $\tilde\bbeta_0=\V\tilde\s_0$ and 
\begin{align*}
    \nabla g_0(\tilde\s_0) 
    &= -\V^{\trans}(L\hat\bbeta_0 - \nabla f(\hat\bbeta_0) - L\V\tilde\s_0)
    = -L\V^{\trans}\V\hat\s_0 + \V^{\trans}\nabla f(\hat\bbeta_0) + L\V^{\trans}\V\tilde\s_0 \\
    &= -L\hat\s_0 + L\tilde\s_0 = \bzero.
\end{align*}
The second equality is obtained be plugging in $\hat\bbeta_0=\V\hat\s_0$ and the last equality is resulted from the fact that the columns of $\V$ are orthogonal basis. Therefore, we get $\hat\s_0=\tilde\s_0$ and it follows that $\hat\bbeta_0 = \tilde\bbeta_0$.

Note that, the constraint $\D\bbeta=0$ corresponds to the case when there is no restriction on the dual parameters. The solution for the dual problem of the surrogate primal problem~\eqref{eq:iniprob2}, $\tilde\u_0$ is used as the initial value for the dual updates, which is 
\begin{align*}
    \tilde\u_0 = \arg\min \frac{1}{2L}\|L\hat\bbeta_0 - \nabla f(\hat\bbeta_0) - \D^{\trans}\u\|^2.
\end{align*}
Here we have replaced $\tilde\bbeta_0$ with $\hat\bbeta_0$. 

Let $\lambda_0=\|\tilde\u_0\|_{\infty}$. Then $(\hat\bbeta_0,\tilde\u_0)$ is a pair of solution to the following problem
\begin{align*}
    \mathop{\mbox{minimize}}_{\bbeta,\z} \frac{1}{2L}\|L\hat\bbeta_0 - \nabla f(\hat\bbeta_0) - L\bbeta\|^2 + \lambda_0\|\z\|_1,\,\mbox{s.t. } \D\bbeta=\z,
\end{align*}
which has the Lagrangian form as 
\begin{align*}
    \mathop{\mbox{minimize}}_{\bbeta,\z, \u} \frac{1}{2L}\|L\hat\bbeta_0 - \nabla f(\hat\bbeta_0) - L\bbeta\|^2 + \lambda_0\|\z\|_1 + \u^{\trans}(\D\bbeta-\z).
\end{align*}
From the stationarity conditions of the above Lagrangian problem at $(\hat\bbeta_0,\tilde\u_0)$ , further we have 
\begin{align}
    \nabla f(\hat\bbeta_0) + \D^{\trans}\tilde\u_0 = \bzero, \label{eq:stationary1} \\
    \lambda_0\hat\bdelta - \tilde\u_0 \ni \bzero,\label{eq:stationary2}
\end{align}
where the $i$th element of $\hat\bdelta$ takes value $\sgn\{(\D\hat\bbeta_0)_i\}$ if $(\D\hat\bbeta_0)_i$ is nonzero, and takes any value in $(-1, 1)$ o.w. Here, $(\D\hat\bbeta_0)_i$ is the $i$th element of $\D\hat\bbeta_0$. By plugging in \eqref{eq:stationary2} into \eqref{eq:stationary1}, we get
\begin{align}\label{eq:stationary3}
    \nabla f(\hat\bbeta_0) + \lambda_0\D^{\trans}\hat\bdelta \ni \bzero,
\end{align}
which is exactly the stationary condition for
\begin{align*}
    \mathop{\mbox{minimize}}\limits_{\bbeta} f(\bbeta) + \lambda_0\|\D\bbeta\|_1.
\end{align*}
It then follows that $\hat\bbeta_0$ is the solution to the generalize lasso problem with tuning parameter $\lambda_0$. This completes the proof. 

Note that, although $\hat\bbeta_0$ is the unique minimizer of the primal problen when $f$ is strictly convex, there can be multiple minimizers for the dual problem when $\D$ is not of full row rank. However, $\D^{\trans}\tilde\u_0$ is unique. By the stationarity condition~\eqref{eq:stationary1}, we have a single $\hat\bbeta_0$ corresponds to several $\tilde\u_0$, and we take $\tilde\u_0$ as the least norm solution when $\D$ does not have full row rank.

\subsection{Proof of Theorem~\ref{thm:thm1}}

%\begin{theorem}\label{thm:thm1}
%If the loss function $f(\bbeta)$ is $\mu$-strongly convex and $L$-smooth, then as the step size $\varepsilon \rightarrow 0$ and when we take $N_m=1$, the MM-DUST path converges to the generalized lasso path uniformly.
%\end{theorem}

Recall that in the MM-DUST path algorithm, we consecutively produce the estimations of $\bbeta$ for a sequence of generalized lasso problems, in which the $t$th problem has a fixed tuning parameter $\lambda_t$, and we have $\lambda_{t} = \lambda_{t-1} - \varepsilon$ with $\varepsilon$ being the fixed step size parameter. Consequently, the total number of points on the solution path equals to $N_{\lambda} = \lambda_{max}/\varepsilon=\lambda_0/\varepsilon$.

Let $\hat\bbeta_t$ be the estimation of $\bbeta$ from MM-DUST when the tuning parameter takes value $\lambda_t$. Consider the original optimization problem and the majorized problem at $\lambda_t$, we have
\begin{align}
    \bbeta_t^* &= \arg\min g_t(\bbeta) = \arg\min f(\bbeta) + \lambda_t\|\D\bbeta\|_1, 
    \label{eq:exactprob}\\
    \tilde\bbeta_t &= \arg\min \tilde g_t(\bbeta|\hat\bbeta_{t-1}) = \arg\min \tilde f(\bbeta|\hat\bbeta_{t-1}) + \lambda_t\|\D\bbeta\|_1, \label{eq:mmprob}
\end{align}
where $g_t$, $f$, $\tilde g_t$, and $\tilde f$ are the objective function, loss function, majorized objective function and majorized loss function, respectively. By design, the $(t-1)$th estimate $\hat\bbeta_{t-1}$ from MM-DUST with tuning parameter $\lambda_{t-1}$ is used as the initial value to construct the surrogate function in the majorization step for the $t$th optimization problem. 

To prove that the MM-DUST path converges to the generalized lasso path uniformly, we will show that
\begin{align*}
	\|\hat\bbeta_t - \bbeta_t^*\| \leq C\varepsilon, \ t = 1, 2, \ldots, N_{\lambda},
\end{align*}
where $C$ is a constant not depending on $\varepsilon$. The above distance between the output from MM-DUST, $\hat\bbeta_t$, and the exact solution $\bbeta_t^*$, can be well-bounded with 
\begin{align}\label{eq:decompose1}
    \|\hat\bbeta_t - \bbeta_t^*\| \leq \|\hat\bbeta_t - \tilde\bbeta_t\| + \|\tilde\bbeta_t - \bbeta_t^*\|.
\end{align}
We will discuss the two parts on the RHS to construct the desired upper bound. 

\subsubsection{Bound $\|\hat\bbeta_t - \tilde\bbeta_t\|$}

In this section, we try to bound the first part on the RHS of \eqref{eq:decompose1}, which measures the distance between the optimal solution of the majorized problem and the output from MM-DUST. 

Based on the exact stationarity condition at the optimal point of the $t$th majorized problem~\eqref{eq:mmprob}, we have
\begin{align*}
    \tilde\bbeta_t &= \hat\bbeta_{t-1} - \frac{1}{L}\{\D^{\trans}\tilde\u_t + \nabla f(\hat\bbeta_{t-1})\},
\end{align*}
where $\tilde\u_t$ is the exact solution of the dual problem~\eqref{eq:mmprob}, which is a convex box-constrained problem with squared loss. Let $\ddot\u_t$ be the 
%From Lemma~\ref{lemma1}, after sufficient iterations from the Dual-Solver, the optimal solution for the dual problem can always be obtained after sufficient iterations from the Dual-Solver. 
%Since the dual problem may have multiple points that satisfy the terminating conditions, we take $\tilde\u_t$ as the 
point that requires the fewest dual updating steps, $N_t$, to satisfy the terminating conditions in Lemma~\ref{lemma:dual}. Further, we take $N_0 = \max_{t} N_t$ for $t=1, 2, \ldots, N_{\lambda}$.

On the other hand, from the updating rule of MM-DUST with at most $N_d$ steps, we also have
\begin{align*}
    \hat\bbeta_t &=  \hat\bbeta_{t-1} - \frac{1}{L}(\D^{\trans}\hat\u_t + \nabla f(\hat\bbeta_{t-1})),
\end{align*}
where $\hat\u_t$ is the output from MM-DUST. As we restrict the maximum number of dual iterations as $N_d$, the result from MM-DUST, $\hat\u_t$, can be different from $\ddot \u_t$. It then follows that 
\begin{align}\label{eq:part1}
    \|\hat\bbeta_t - \tilde\bbeta_t\| 
    &= \frac{1}{L}\|\D(\hat\u_t - \tilde\u_t)\| 
    = \frac{1}{L}\|\D(\hat\u_t - \ddot\u_t + \ddot\u_t - \tilde\u_t)\| \notag\\
    &\leq \frac{1}{L}(\|\D\|_F\|\hat\u_t - \ddot\u_t\| + \sqrt{L_1\lambda_t m\varepsilon}) \notag\\
    &\leq \frac{1}{L}(\|\D\|_F|N_0-N_d|\sqrt{m}\varepsilon + \sqrt{L_1\lambda_0m\varepsilon}).
\end{align}
The first inequality is derived based on the results in Appendix~\ref{supp:dual}, where we derive the suboptimality bound for $\ddot\u_t$ and $L_1$ is the smoothness parameter for the dual problem.
The last inequality follows from the fact that in each of the dual updates, at most $m$ elements of $\u$ are updated by the fixed step size $\varepsilon$.

%=========================================================
\subsubsection{Bound $\|\tilde\bbeta_t - \bbeta_t^*\|$}

Next, we try to bound the second part on the RHS in \eqref{eq:decompose1}, which measures the distance between the optimal point of the exact objective function $g_t$ and the majorized objective function$\tilde g_t$. Assume that the loss function $f$ is $\mu$-strongly convex, $L$-smooth, and twice-differentiable. %We first show that the objective function $g$ is also $\mu$-strongly convex.
The upper bound is then constructed by discussing the following three gaps: $g_t(\tilde\bbeta_t)-g_t(\bbeta_t^*)$, $\tilde g_t(\tilde\bbeta_t)-\tilde g_t(\bbeta_t^*)$, and $\tilde g_t(\bbeta_t^*)-g_t(\bbeta_t^*)$.

Firstly, we quantify $g_t(\tilde\bbeta_t)-g_t(\bbeta_t^*)$. Based on the $\mu$-strongly convexity of $f$, for $\forall \bbeta_1$, $\bbeta_2\in\mathcal{R}^p$, we have
\begin{align*}
    f(\bbeta_1) \geq f(\bbeta_2) + \nabla f(\bbeta_2)^{\trans}(\bbeta_1 - \bbeta_2) + \frac{\mu}{2}\|\bbeta_1-\bbeta_2\|^2.
\end{align*}
Since the penalty function is also convex, we also have 
\begin{align*}
    \lambda\|\D\bbeta_1\|_1 \geq \lambda\|\D\bbeta_2\|_1 + \lambda(\D^{\trans}\bdelta_2)^{\trans}(\bbeta_1 -\bbeta_2), \forall \lambda>0,
\end{align*}
where $\bdelta_2\in\mathcal{R}^m$ takes value $\sgn(\D\bbeta_2)_i$ at the $i$th position. Therefore, we can get 
\begin{align}\label{eq:stronglyconvex}
    g(\bbeta_1) \geq g(\bbeta_2) + (\nabla f(\bbeta_2) + \lambda\D^{\trans}\bdelta_2)^{\trans}(\bbeta_1-\bbeta_2) + \frac{\mu}{2}\|\bbeta_1 - \bbeta_2\|^2,
\end{align}
where $\nabla f(\bbeta_2) + \lambda\D^{\trans}\bdelta_2 = \partial g(\bbeta_2)$ is the subgradient of the objective function $g$ evaluated at $\bbeta_2$. 
%It indicates that $g$ is also $\mu$-strongly convex. 
Set $\bbeta_1=\tilde\bbeta_t$ and $\bbeta_2=\bbeta_t^*$. Since $\bbeta_t^*$ is the minimum point of $g_t$, we have $\bzero \in \nabla f(\bbeta_t^*) + \lambda_t\D^{\trans}\bdelta_t$. It indicates that
\begin{align}\label{eq:bound1}
    g_t(\tilde\bbeta_t) \geq g_t(\bbeta_t^*) + \frac{\mu}{2}\|\bbeta_t^* - \tilde\bbeta_t\|^2.
\end{align}

Next, we discuss the gap $\tilde g_t(\tilde\bbeta_t|\hat\bbeta_{t-1})-\tilde g_t(\bbeta_t^*|\hat\bbeta_{t-1})$. Recall that the second order gradient of the majorized loss function $\tilde f(\bbeta|\hat\bbeta_{t=1})$ equals to $L\I$, which indicates that the majorized loss is $L$-strongly convex. The penalty part in the majorized problem is the same as that in the original problem. By the same arguments used to show \eqref{eq:stronglyconvex}, it is straightforward to verify that %$\tilde g_t(\bbeta|\hat\bbeta_{t-1})$ is $L$-strongly convex. Thus, we can get
\begin{align*}
    \tilde g_t(\bbeta_t^*|\hat\bbeta_{t-1}) \geq \tilde g_t(\tilde \bbeta_t|\hat\bbeta_{t-1}) + 
    \partial\tilde g_t(\tilde\bbeta_t|\hat\bbeta_{t-1})^{\trans}(\bbeta_t^* - \tilde\bbeta_t) + \frac{L}{2}\|\bbeta_t^* - \tilde\bbeta_t\|^2.
\end{align*}
As $\tilde\bbeta_t \in \arg\min \tilde g_t(\bbeta|\hat\bbeta_{t-1})$, we have $\bzero \in \partial\tilde g_t(\tilde\bbeta_t|\hat\bbeta_{t-1})$. As a consequence, 
\begin{align}\label{eq:bound2}
    \tilde g_t(\bbeta_t^*|\hat\bbeta_{t-1}) \geq \tilde g_t(\tilde \bbeta_t|\hat\bbeta_{t-1}) + 
     \frac{L}{2}\|\bbeta_t^* - \tilde\bbeta_t\|^2.
\end{align}

Finally, we try to bound $\tilde g_t(\bbeta_t^*|\hat\bbeta_{t-1})-g_t(\bbeta_t^*)$, which measures the gap between the original and majorized objective function at the optimal of the original problem. Recall that in the MM-DUST algorithm, we take the majorized objective function as 
\begin{align*}
    \tilde g_t(\bbeta|\hat\bbeta_{t-1}) = f(\hat\bbeta_{t-1}) + \nabla f(\hat\bbeta_{t-1})^{\trans}(\bbeta - \hat\bbeta_{t-1}) + 
    \frac{L}{2}\|\bbeta - \hat\bbeta_{t-1}\|^2 + \lambda_t\|\D\bbeta\|_1
\end{align*}
with $L$ being the Lipschitz continuous constant of the loss function $f$. 
Define the gap $h_t(\bbeta) = \tilde g_t(\bbeta|\hat\bbeta_{t-1}) - g_t(\bbeta)$. Then
\begin{align*}
    h_t(\bbeta) 
    &= f(\hat\bbeta_{t-1}) - f(\bbeta) + \nabla f(\hat\bbeta_{t-1})^{\trans}(\bbeta - \hat\bbeta_{t-1}) + 
    \frac{L}{2}\|\bbeta - \hat\bbeta_{t-1}\|^2, \\
    \nabla h_t(\bbeta) &= -\nabla f(\bbeta) + \nabla f(\hat\bbeta_{t-1}) + L(\bbeta-\hat\bbeta_{t-1}), \\
    \nabla^2 h_t(\bbeta) &= L\I - \nabla^2 f(\bbeta).
\end{align*}
Since both $L\I - \nabla^2 f(\bbeta)$ and $L\I -(L\I - \nabla^2 f(\bbeta))$ are positive semi-definite for all $\bbeta$, the function $h_t$ is convex and $L$-smooth. Consequently, we get
\begin{align*}
    h_t(\bbeta_t^*) &\leq h_t(\hat\bbeta_{t-1}) + \nabla h_t(\hat\bbeta_{t-1})^{\trans}(\bbeta_t^* - \hat\bbeta_{t-1}) + \frac{L}{2}\|\bbeta_t^* - \hat\bbeta_{t-1}\|^2.
\end{align*}
As the majorization is taken at $\hat\bbeta_{t-1}$, we have $h_t(\hat\bbeta_{t-1})=0$ and $\nabla h_t(\hat\bbeta_{t-1})=\bzero$. Thus, we get 
\begin{align}\label{eq:bound3}
    |h_t(\bbeta_t^*)| &\leq \frac{L}{2}\|\bbeta_t^* - \hat\bbeta_{t-1}\|^2.
\end{align}

According to the property of the majorization step, we also have $g_t(\tilde\bbeta_t) \leq \tilde g_t(\tilde\bbeta_t|\hat\bbeta_{t-1})$. Combining this property with  inequalities \eqref{eq:bound1},  \eqref{eq:bound2}, and \eqref{eq:bound3}, we get
\begin{align*}
    g_t(\tilde\bbeta_t) + \frac{L}{2}\|\bbeta_t^* - \tilde\bbeta_t\|^2
    &\leq \tilde g_t(\tilde\bbeta_t|\hat\bbeta_{t-1}) + \frac{L}{2}\|\bbeta_t^* - \tilde\bbeta_t\|^2 \\
    \mbox{by}\ \eqref{eq:bound2} &\leq \tilde g_t(\bbeta_t^*|\hat\bbeta_{t-1}) \\
    &= g_t(\bbeta_t^*) + h(\bbeta_t^*) \\
    \mbox{by}\ \eqref{eq:bound1}, \eqref{eq:bound3}&\leq g_t(\tilde\bbeta_t) - \frac{\mu}{2}\|\bbeta_t^* - \tilde\bbeta_t\|^2 + \frac{L}{2}\|\bbeta_t^* - \hat\bbeta_{t-1}\|^2.
\end{align*}
That is 
\begin{align}\label{eq:part2}
    \|\bbeta_t^* - \tilde\bbeta_t\|^2 \leq \frac{L}{L + \mu}\|\bbeta_t^* - \hat\bbeta_{t-1}\|^2.
\end{align}
Since $L/(L+\mu)$ is smaller than $1$, inequality ~\eqref{eq:part2} indicates that for the $t$th optimization problem, the optimal solution $\tilde\bbeta_t$ from the one-time majorization-minimization is always improved when compared with the initial estimate $\hat\bbeta_{t-1}$.

%===========================================================
\subsubsection{Bound $\|\hat\bbeta_t - \bbeta_t^*\|$}

Combining \eqref{eq:decompose1}, \eqref{eq:part1} and \eqref{eq:part2}, we get 
\begin{align}
    \|\hat\bbeta_t - \bbeta_t^*\| \leq \frac{1}{L}(\|\D\|_F|N_0-N_d|\sqrt{m}\varepsilon  + \sqrt{L_1\lambda_0m\varepsilon})
    + \sqrt{\frac{L}{L+\mu}}\|\bbeta_t^* - \hat\bbeta_{t-1}\|.
\end{align}
Further, we have $\|\bbeta_t^* - \hat\bbeta_{t-1}\| \leq \|\bbeta_t^* - \bbeta_{t-1}^*\| + \|\bbeta_{t-1}^* - \hat\bbeta_{t-1}\|$. The above result indicates that the final bound can be derived by induction. 

Next, we try to bound $\|\bbeta_{t}^* - \bbeta_{t-1}^*\|$ by a function of the step size $\varepsilon$. Still from the $\mu$-strongly convexity of the loss function $f$, by setting $\bbeta_1 = \bbeta_t^*$ and $\bbeta_2 = \bbeta_{t-1}^*$ in \eqref{eq:stronglyconvex}, we have
\begin{align*}
    g_t(\bbeta_t^*) \geq g_t(\bbeta_{t-1}^*) + (\nabla f(\bbeta_{t-1}^*) + \lambda_t\D^{\trans}\bdelta_{t-1})^{\trans}(\bbeta_t^*-\bbeta_{t-1}^*) + \frac{\mu}{2}\|\bbeta_t^* - \bbeta_{t-1}^*\|^2,
\end{align*}
where the subgradient $\bdelta_{t-1}$ is evaluated at $\bbeta_{t-1}^*$. Since $\bbeta_t^*$ is the minima, it holds that $g_t(\bbeta_t^*) \leq g_t(\bbeta_{t-1}^*)$. By applying the Cauchy-Schwartz inequality, we get
\begin{align*}
    0 \geq -\|\nabla f(\bbeta_{t-1}^*) + \lambda_t\D^{\trans}\bdelta_{t-1}\| \|\bbeta_t^* - \bbeta_{t-1}^*\| + \frac{\mu}{2}\|\bbeta_t^* - \bbeta_{t-1}^*\|^2.
\end{align*}

It then follows that 
\begin{align}\label{eq:thirdeq}
    \|\bbeta_t^* - \bbeta_{t-1}^*\| 
    & \leq \frac{2}{\mu}\|\nabla f(\bbeta_{t-1}^*) + \lambda_t\D^{\trans}\bdelta_{t-1}\| \notag\\
    & \leq \frac{2}{\mu}\|\nabla f(\bbeta_{t-1}^*) + (\lambda_t+\varepsilon)\D^{\trans}\bdelta_{t-1} - \varepsilon\D^{\trans}\bdelta_{t-1}\| \notag\\
    & \leq \frac{2}{\mu}\|\varepsilon\D^{\trans}\bdelta_{t-1}\| \notag\\
    & \leq \frac{2}{\mu}\varepsilon \|\D\|_F\sqrt{m}.
\end{align}
The third inequality follows from the fact that  $\lambda_t + \varepsilon = \lambda_{t-1}$ and $\bzero \in \nabla f(\bbeta_{t-1}^*) + \lambda_{t-1}\D^{\trans}\bdelta_{t-1} = \partial g_{t-1}(\bbeta_{t-1}^*)$, as $\bbeta_{t-1}^*$ is the mimia of the optimization problem at $\lambda_{t-1}$. 

Let $c = \sqrt{L/(L + \mu)} < 1$ and $c_0 = 1/L\|\D\|_F|N_t - N_d| \sqrt{m}\varepsilon + 
	2c/\mu\varepsilon \|\D\|_F\sqrt{m}$. 
Finally, 
\begin{align*}
	\|\hat\bbeta_t - \bbeta_t^*\| 
	\leq &\ \frac{1}{L}(\|\D\|_F|N_0-N_d|\sqrt{m}\varepsilon + \sqrt{L_1\lambda_0m\varepsilon}) + 
	c\|\bbeta_t^* - \hat\bbeta_{t-1}\| \\
	%\leq &\ \varepsilon\sqrt{m}\|\D\|_F(\frac{1}{L}|N_0 - N_d|  +  \frac{2c}{\mu}) + c \|\hat\bbeta_{t-1} - \bbeta_{t-1}^*\| \\
    \leq &\  \frac{1}{L}(\|\D\|_F|N_0-N_d|\sqrt{m}\varepsilon + \sqrt{L_1\lambda_0m\varepsilon}) + \frac{2c}{\mu}\varepsilon \|\D\|_F\sqrt{m}  +
	c \|\hat\bbeta_{t-1} - \bbeta_{t-1}^*\| \\
    = &\ A + c\|\hat\bbeta_{t-1} - \bbeta_{t-1}^*\|\\
	%& = \varepsilon\sqrt{m}\|\D\|_F(\frac{1}{L}|N_t - N_d|  + 
	%\frac{2c}{\mu} + c\|\hat\bbeta_{t-1} - \bbeta_{t-1}^*\| \\
	%\leq &\ \varepsilon\sqrt{m}\|\D\|_F(\frac{1}{L}|N_0 - N_d|  + 
	%\frac{2c}{\mu}) + c\left( \varepsilon\sqrt{m}\|\D\|_F(\frac{1}{L}|N_0 - N_d|  + 
	%\frac{2c}{\mu}) + c\|\hat\bbeta_{t-2} - \bbeta_{t-2}^*\|\right) \\
    \leq &\ A + c(A + c\|\hat\bbeta_{t-2} - \bbeta_{t-2}^*\|)\\
	%\leq &\ \varepsilon\sqrt{m}\|\D\|_F(\frac{1}{L}|N_0 - N_d|  + 
	%\frac{2c}{\mu})(1 + c + c^2 + \ldots, + c^{t-1}) + c^{t}\|\hat\bbeta_0 - \bbeta_0^*\| \\
    \leq &\ A(1 + c + c^2 + \ldots, + c^{t-1}) + c^{t}\|\hat\bbeta_0 - \bbeta_0^*\| \\
%    \leq &\ \varepsilon\sqrt{m}\|\D\|_F\frac{1}{L}|N_0 - N_d|(1 + c + c^2 + \ldots, + c^{t-1}) 
%    + \varepsilon\sqrt{m}\|\D\|_F\frac{2c}{\mu}(1 + c + c^2 + \ldots, + c^{t-2}) \\
%    &+ c^t\|\bbeta_1^* - \hat\bbeta_0\| +  c^t\|\hat\bbeta_0 - \bbeta_0^*\| \\
%    = &\ \varepsilon\sqrt{m}\|\D\|_F\frac{1}{L}|N_0 - N_d|\frac{1-c^t}{1-c} + 
%    \varepsilon\sqrt{m}\|\D\|_F\frac{2c}{\mu}\frac{1-c^{t-1}}{1-c} + 
%    c^t\|\bbeta_1^* - \hat\bbeta_0\| +  c^t\|\hat\bbeta_0 - \bbeta_0^*\| \\
    \leq &\ \frac{1}{1-c} A+ c^t\|\hat\bbeta_0 - \bbeta_0^*\|\\
    = &\ \frac{1}{1-c}(\frac{1}{L}\|\D\|_F|N_0-N_d|\sqrt{m}\varepsilon + \frac{1}{L}\sqrt{L_1\lambda_0m\varepsilon}+ \frac{2c}{\mu}\varepsilon \|\D\|_F\sqrt{m})+ c^t\|\hat\bbeta_0 - \bbeta_0^*\|.\\
	%\leq &\ \frac{1}{1-c} \varepsilon\sqrt{m}\|\D\|_F(\frac{1}{L}|N_0- N_d|  + 
	%\frac{2c}{\mu})+ c^t\|\hat\bbeta_0 - \bbeta_0^*\|.
\end{align*}

Recall that in Algorithm~\ref{alg:MM-DUST}, we round the first dual estimate $\hat\u_0$ and obtain the first tuning parameter $\lambda_0$, and $\bbeta_0^*$ is the exact primal solution corresponding to $\lambda_0$. Let $\ddot\lambda_0$ be the infinity norm of $\hat\u_0$ before rounding and $\ddot\bbeta_0^*$ be the exact solution corresponding to $\ddot\lambda_0$. We take $\hat\bbeta_0$ as the first estimate from MM-DUST, which is also the initial value for the majorization step when $\lambda=\lambda_0-\varepsilon$. By Lemma~\ref{lemma2}, we have $\hat\bbeta_0=\ddot\bbeta_0^*$.

As we round $\hat\u_0$ by $\varepsilon$, we have $|\lambda_0 - \ddot\lambda_0|<\varepsilon$. By same argument as inequality~\eqref{eq:thirdeq}, we have
\begin{align*}
    \|\hat\bbeta_0 - \bbeta_0^*\| = \|\ddot\bbeta_0^* - \bbeta_0^*\| \leq \frac{2}{\mu}\varepsilon \|\D\|_F\sqrt{m}.
\end{align*}

%By design of the algorithm, Lemma~\ref{lemma2} guarantees that $\hat\bbeta_0=\bbeta_0^*$. 
It then follows that $\|\hat\bbeta_t - \bbeta_t^*\|  \rightarrow 0$ as $\varepsilon \rightarrow 0$.
%as long as
%\begin{align*}
%	\frac{1}{1-c} \varepsilon\sqrt{m}\|\D\|_F(\frac{1}{L}|N_0- N_d|  + 
%	\frac{2c}{\mu}) = \mathcal{O}(\varepsilon).
%\end{align*}
This completes the proof.

\section{Additional Materials for Simulations}
\label{sec:app:simu}

\subsection{Additional Results for Section~\ref{sec:converge-logit} and \ref{sec:converge-cox}}
\label{sec:app:fullpath}

\begin{figure}[t]
     \centering
     \begin{subfigure}[b]{0.24\textwidth}
         \centering
         \includegraphics[width=\textwidth, height=\textwidth]{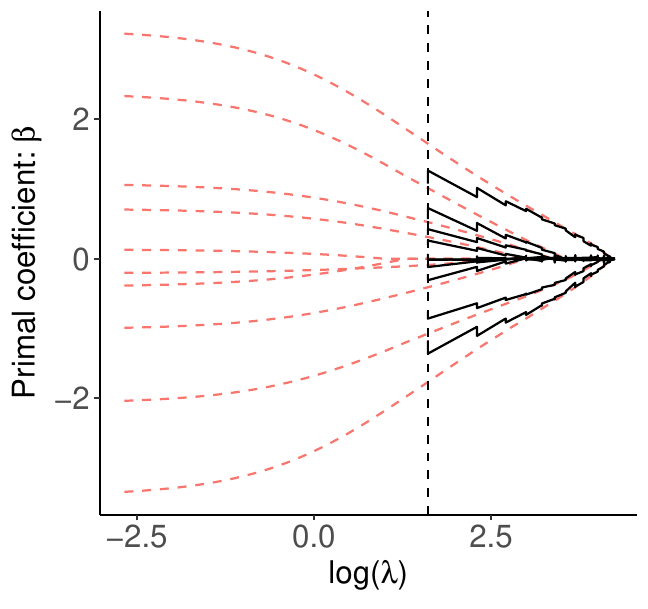}
         %\caption{$\varepsilon=5$}
     \end{subfigure}
     \hfill
     \begin{subfigure}[b]{0.24\textwidth}
         \centering
         \includegraphics[width=\textwidth, height=\textwidth]{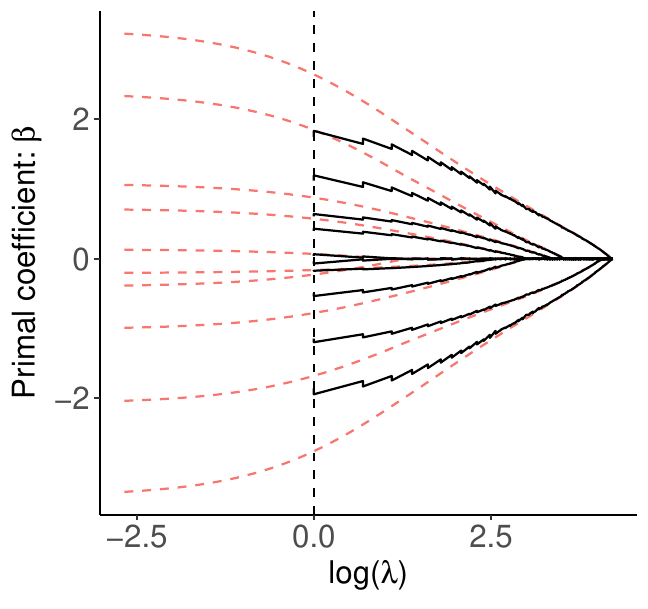}
         %\caption{$\varepsilon=1$}
      \end{subfigure}
     \begin{subfigure}[b]{0.24\textwidth}
         \centering
         \includegraphics[width=\textwidth, height=\textwidth]{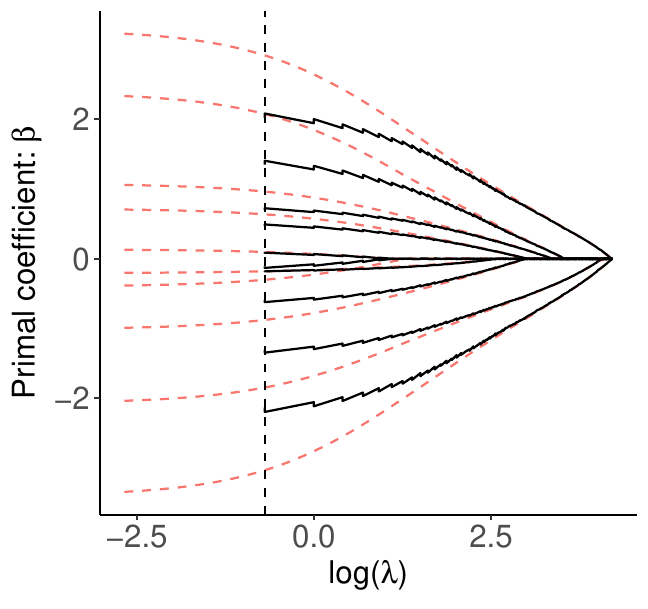}
         %\caption{$\varepsilon=0.5$}
      \end{subfigure}
      \hfill
     \begin{subfigure}[b]{0.24\textwidth}
         \centering
         \includegraphics[width=\textwidth, height=\textwidth]{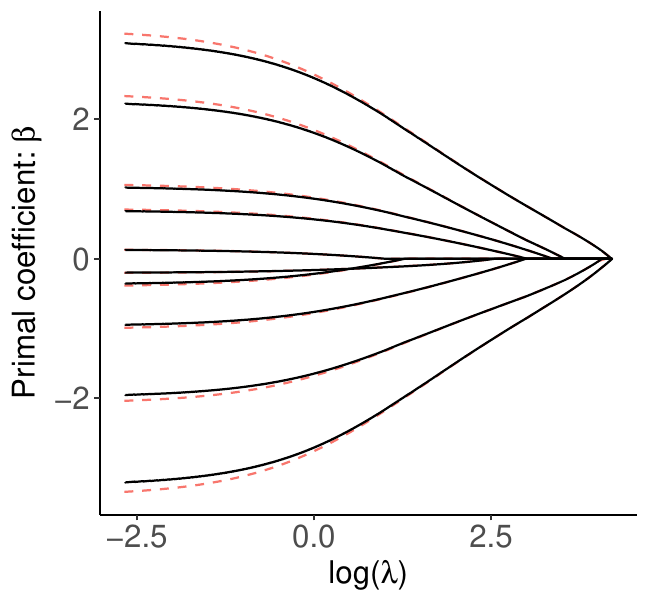}
         %\caption{$\varepsilon=0.01$}
      \end{subfigure}

      \begin{subfigure}[b]{0.24\textwidth}
         \centering
         \includegraphics[width=\textwidth, height=\textwidth]{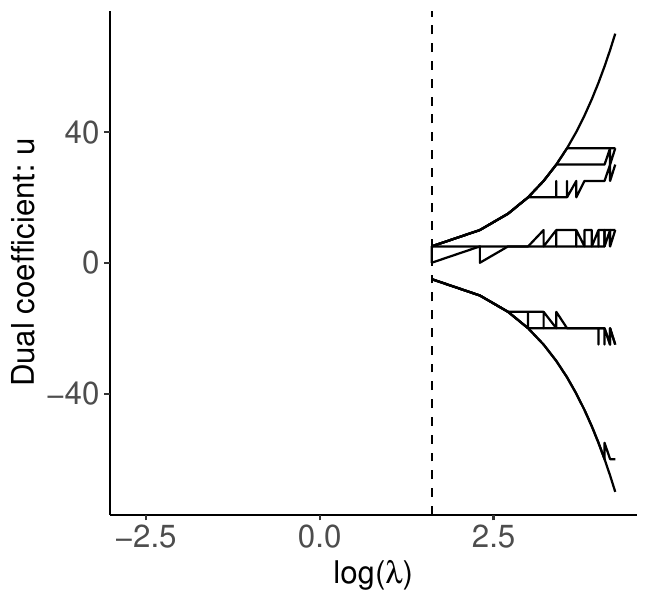}
         \caption{$\varepsilon=5$}
     \end{subfigure}
     \hfill
     \begin{subfigure}[b]{0.24\textwidth}
         \centering
         \includegraphics[width=\textwidth, height=\textwidth]{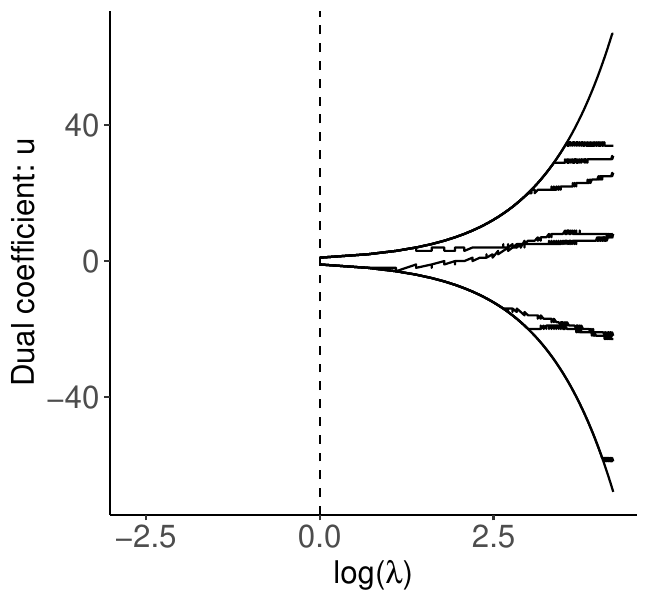}
         \caption{$\varepsilon=1$}
      \end{subfigure}
     \begin{subfigure}[b]{0.24\textwidth}
         \centering
         \includegraphics[width=\textwidth, height=\textwidth]{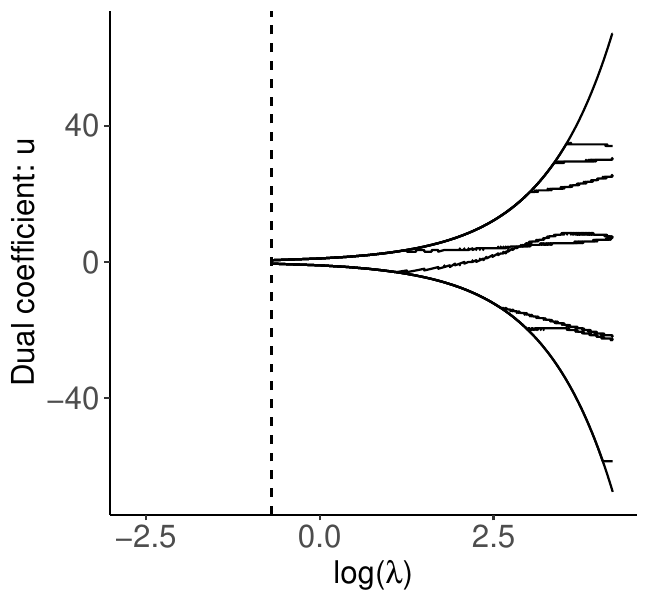}
         \caption{$\varepsilon=0.5$}
      \end{subfigure}
      \hfill
     \begin{subfigure}[b]{0.24\textwidth}
         \centering
         \includegraphics[width=\textwidth, height=\textwidth]{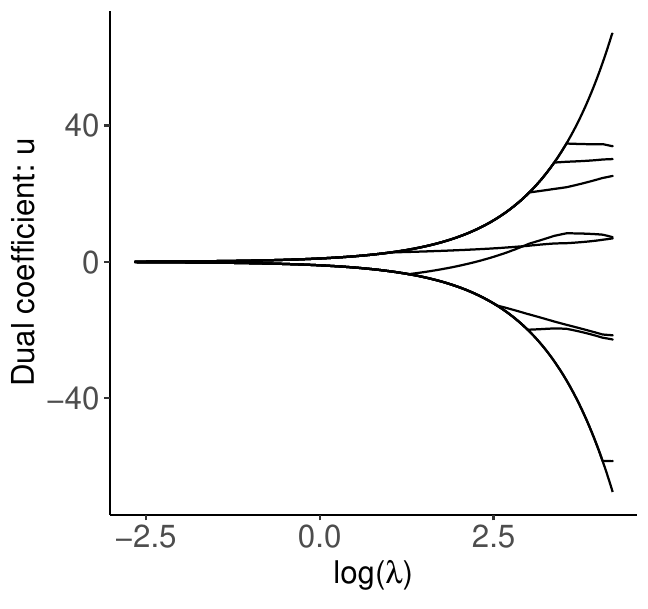}
         \caption{$\varepsilon=0.01$}
      \end{subfigure}

      \caption{Simulation: full solution paths of $\bbeta$ and $\u$ with different step sizes. The exact solution paths from \texttt{glmnet} are shown in red dashed lines, while the paths from MM-DUST are shown in black solid lines. The top four figures are the paths for the primal coefficient $\bbeta$, with the x-axis as $\log(\lambda)$. The bottom four figures are the paths for the dual coefficient $\u$, with the x-axis as $\log(\lambda)$. The vertical dashed line marks the point when $\|\hat{\u}\|_{\infty}\leq \varepsilon$ and the algorithm stops.
      }
  \label{fig:path-logit-full}
\end{figure}

We provide the full solution paths for the example in Section~\ref{sec:converge-logit} in Figure~\ref{fig:path-logit-full}, in which the range of the x-axis is determined by the range of $\lambda$ sequence provided by the \textit{glmnet} package.

\begin{figure}[ht]
    \centering
    \includegraphics[width=\textwidth]{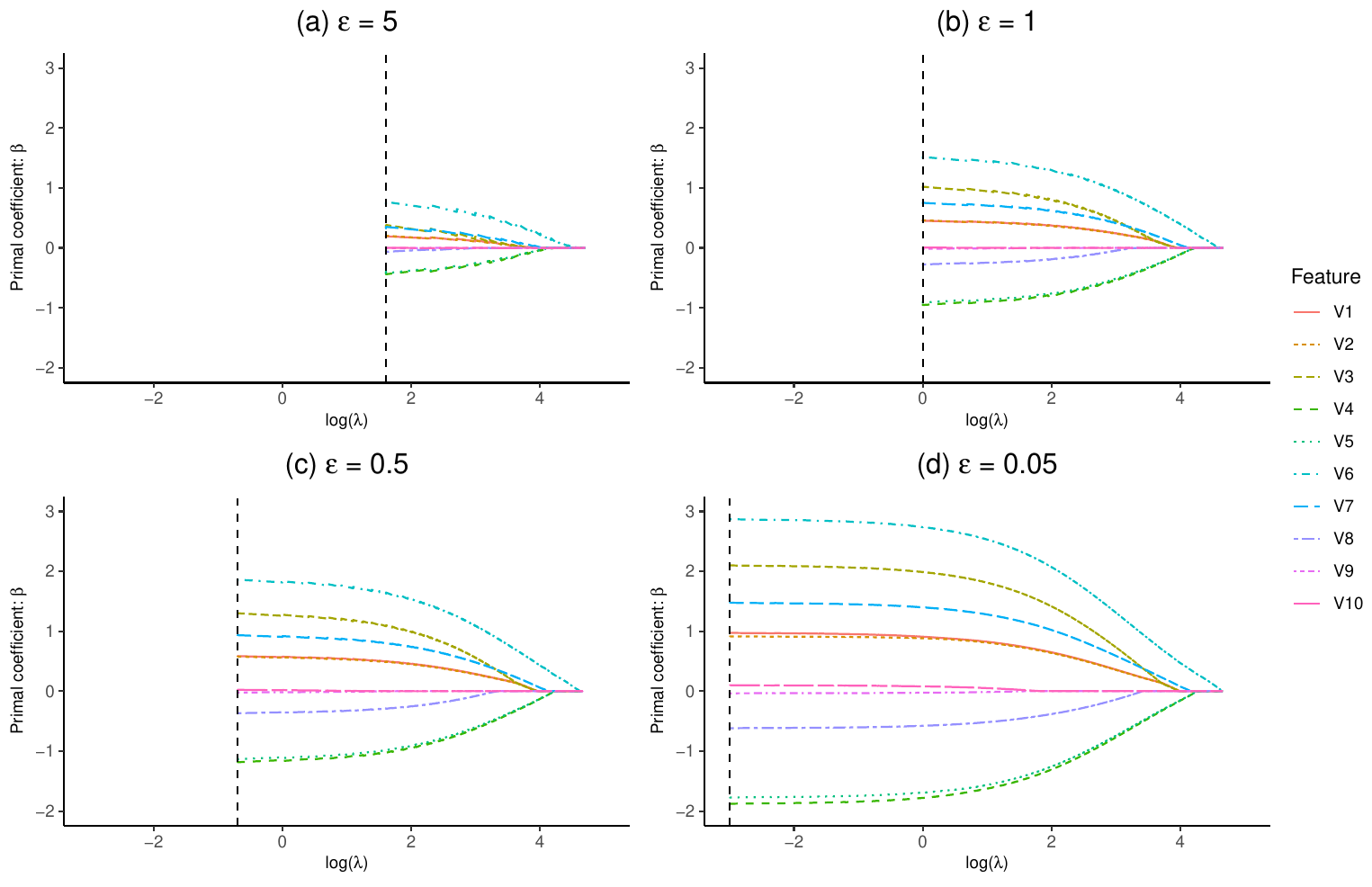}
    \caption{Simulation: full primal solution paths of $\bbeta$ for different step sizes.}
  \label{fig:path-cox-full}
\end{figure}

\begin{figure}[ht]
    \centering
    \includegraphics[width=\textwidth]{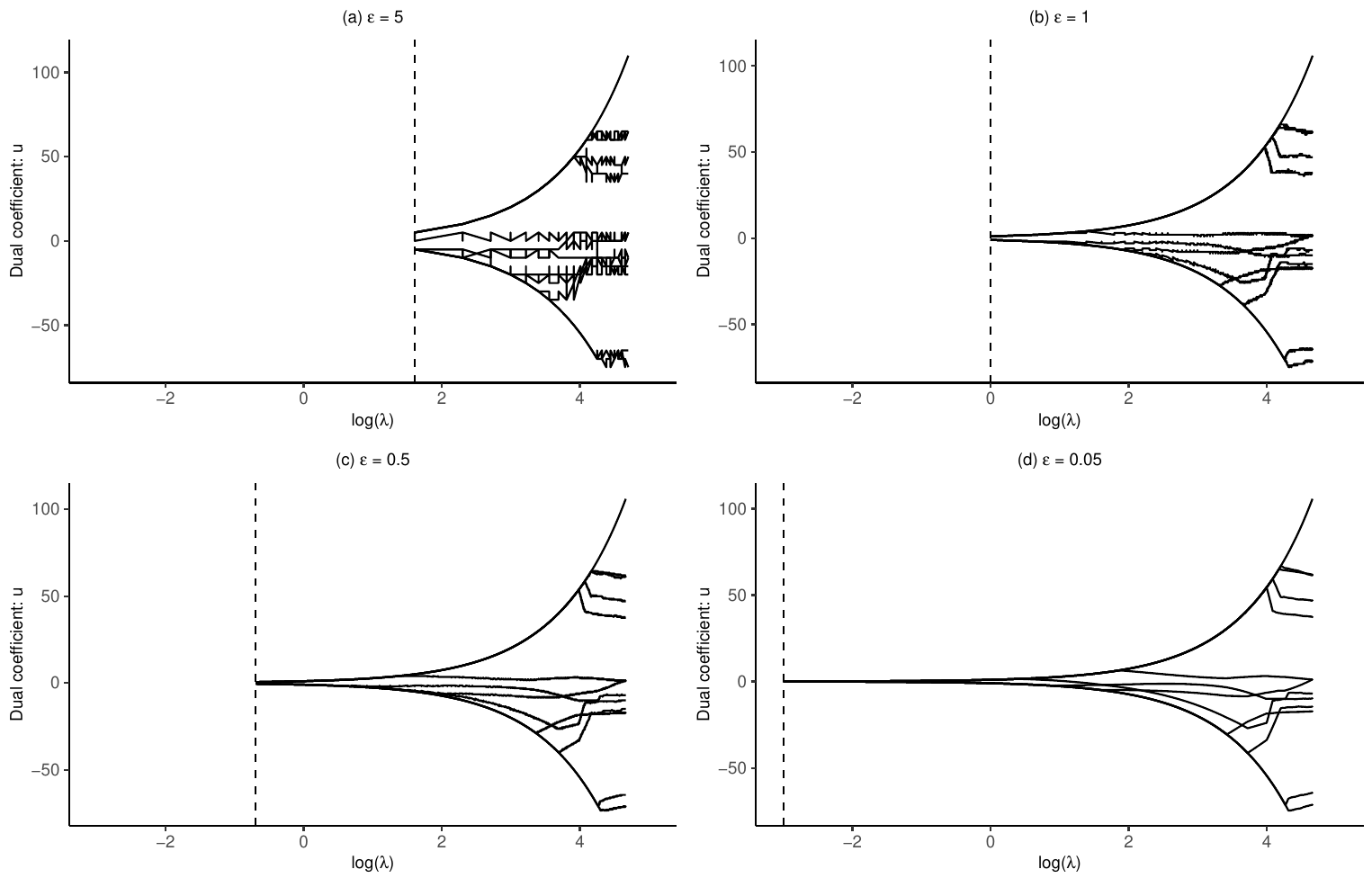}
    \caption{Simulation: full dual solution paths of $\u$ for different step sizes.}
  \label{fig:path-cox-dual-full}
\end{figure}

We provide the full solution paths the example in Section~\ref{sec:converge-cox} in Figure~\ref{fig:path-cox-full} and \ref{fig:path-cox-dual-full}.

\subsection{A Generalized Lasso Problem for Feature Aggregation}

We consider the generalized lasso problem proposed in \citet{yan2021rare}, which is developed to aggregate a large number of features into a relatively smaller number of features that provide better predictions to the response under the guidance of a known tree structure. To be self-included, we briefly introduce the feature aggregation method, and discuss how to apply our algorithm on the optimization.

\begin{figure}
    \centering
    \includegraphics[width=0.5\linewidth]{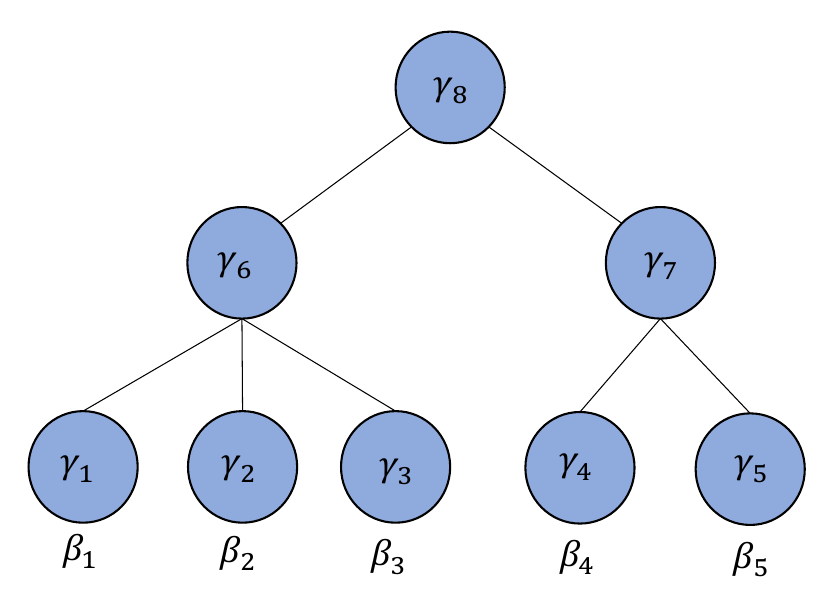}
    \caption{Tree-guided reparameterization. This is a case with $p_0=5$ original features and each circle stands for a node in the tree. For example, we take $\beta_1=\gamma_1+\gamma_6+\gamma_8$. }
    \label{fig:tree}
\end{figure}

Under the settings of linear regression, two original predictors $x_1$ and $x_2$ can be aggregated by summation, if their regression coefficients $\beta_1$ and $\beta_2$ satisfy the following equality constraint:
\begin{align}\label{eq:treeeql}
    \beta_1 = \beta_2.
\end{align}
Given an original design matrix $\X\in\mathbb{R}^{n\times p_0}$ and a tree structure where all the original features are at the leaf nodes,  a reparameterization step is performed to examine the equality constraints and conduct aggregation. An intermediate parameter $\gamma_u$ is assigned to each node $u$ in the tree, and the regression coefficient for the $j$th original predictor $\beta_j$ can be expressed as 
\begin{align*}
    \beta_j = \sum_{u\in ancestor(j)\cup \{\j\}}\gamma_u,
\end{align*}
where $ancestor(j)$ stands for ancerstor node set of the $j$th leaf node. An illustration is shown in Figure~\ref{fig:tree}. Let $\bgamma=(\bgamma_u)\in\mathbb{R}^{|T|}$ be the vector collecting all the intermediate coefficients with $|T|$ being the number of nodes in the given tree. The transformation from $\bbeta$ to $\bgamma$ is linear and can be expressed as $\bbeta=\A\bgamma$ with $\A$ of dimension $p_0\times |T|$.

With the reparameterization step, it can be verified that the equal-sparsity on $\bbeta$ can be converted to zero-sparsity on $\bgamma$. To pursue feature aggregation, the following optimization problem is considered: 
\begin{align*}
    (\hat\bbeta,\hat\bgamma) \in \mathop{\arg\min}_{\bbeta,\bgamma} f(\X\bbeta) + \lambda(\alpha\|\bgamma\|_1 + (1-\alpha)\|\bbeta\|_1),\, \mbox{s.t. } \bbeta = \A\bgamma,
\end{align*}
where $f(\cdot)$ is a loss function depending on the response. As the sparsity in the original features is also desirable, an $\ell_1$ penalty is applied on the $\bbeta$ coefficient. 
When taking $\alpha = 1/2$ and plugging in the constraint $\bbeta=\A\bgamma$, the above optimization problem can be considered as a generalized lasso problem with respect to $\bgamma$
\begin{align*}
    \hat\bgamma  
    & \in\mathop{\arg\min}_{\bgamma} f(\X\A\bgamma) + \lambda(\|\bgamma\|_1 + \|\A\bbeta\|_1), \\
    & \in\mathop{\arg\min}_{\bgamma} f(\X\A\bgamma) + \lambda\|\D\bgamma\|_1,
\end{align*}
where $D = (\I_{|T|}, \A^{\trans})^{\trans} \in \mathbb{R}^{(p_0+|T|)\times|T|}$ and it is of full column rank. We can then apply the MM-DUST algorithm to solve for the $\bgamma$ coefficient. 

\subsection{Data Generation for Simulation Studies}

Without loss of generality, we assume there are no unpenalized covariates. Based on the synthetic tree structure in Figure~\ref{fig:simu_tree_cox}, we have $p_0=42$ original features in the design matrix $\X$. For each row $\x$ in the design matrix $\X$, we have $\x\sim N_{p_0}(\bzero,\Gamma)$, where $\Gamma=(r_{ij})_{p_0\times p_0}$ with $r_{ij}=0.5^{|i-j|}$. The simulations are conducted under the Cox regression settings. We generate the true event time as $T=\exp(\eta + \varepsilon)$, where 
\begin{align*}
\eta = (x_1 + \cdots + x_{12}) - 2(x_{13}+\cdots + x_{18}) 
+ 1.5(x_{19}+ x_{20}+ x_{21}) - 1.5x_{22} - 3x_{23} + 3x_{25},    
\end{align*}
and $\varepsilon\sim N(0, \sigma^2)$ with $\sigma^2$ determined by the desired signal-to-noise ratio (SNR) defined as $\var({\eta})/\sigma^2$. The censoring time $C$ is generated from an exponential distribution with rate parameter 5000. The final survival time of the $i$th observation is then given as $y_i=\min(T_i, C_i)$ with no ties in the simulation studies. 

In this feature aggregation problem, we apply the MM-DUST algorithm to on the $\bgamma$ parameter. Since the tree in Figure~\ref{fig:simu_tree_cox} has $p_0=42$ leaf nodes and $25$ internal nodes with $|T|=67$, the $\D$ matrix is of dimension $109\times 67$ for simulations in Section~\ref{sec:glasso-simu-snr}. As for simulations in Section~\ref{sec:glasso-simu-highd}, the true response $y$ is generated from the same process. We add more nodes to the tree structure and increase the complexity of the structure, but the corresponding added features do not contribute to the true model.

%With a known tree structure, we only consider summing features within the same branch to form the new aggregated features. As a result, a tree with $m$ internal nodes can give $m$ aggregated features. A binary matrix $\A_1\in\mathbb{R}^{p\times m}$ is constructed based on the given tree structure, such that each column of the $\A_1$ matrix is an aggregation rule. The augmented design matrix containing both the original features and the aggregated features can be formulated as $\tilde\X=\X\A\in \mathbb{R}^{n\times(p+m)}$, where $\A=(\I_p,\A_1)$. 

\subsection{Tree Structure}
\label{sec:app:tree}

\begin{figure}[H]
        \centering
        \begin{tikzpicture}[grow = right, level distance = 60pt]
        \tiny
            \tikzstyle{level 1}=[sibling distance=45mm]
            \tikzstyle{level 2}=[sibling distance=21mm]
            \tikzstyle{level 3}=[sibling distance=9.3mm]
            \tikzstyle{level 4}=[sibling distance=3.2mm]
            \node {$u_1^0$} [align=center]
                child {node {$u_1^1$}
                    child {node {$u_1^2$}
                        child {node {$u_1^3$}
                            child {node {$u_1^4\ x_1$}}
                            child {node {$u_2^4\ x_2$}}
                            child {node {$u_3^4\ x3$}}}
                        child {node {$u_2^3$}
                            child {node {$u_4^4\ x_4$}}
                            child {node {$u_5^4\ x_5$}}
                            child {node {$u_6^4\ x_6$}}}
                        }
                    child {node {$u_2^2$}
                        child {node {$u_3^3$}
                            child {node {$u_7^4\ x_7$}}
                            child {node {$u_8^4\ x_8$}}
                            child {node {$u_9^4\ x_9$}}
                            }
                        child {node {$u_4^3$}
                            child {node {$u_{10}^4\ x_{10}$}}
                            child {node {$u_{11}^4\ x_{11}$}}
                            child {node {$u_{12}^4\ x_{12}$}}}}
                    }
                child {node {$u_2^1$}
                    child {node {$u_3^2$}
                        child {node {$u_5^3$}
                            child {node {$u_{13}^4\ x_{13}$}}
                            child {node {$u_{14}^4\ x_{14}$}}
                            child {node {$u_{15}^4\ x_{15}$}}}
                        child {node {$u_6^3$}
                            child {node {$u_{16}^4\ x_{16}$}}
                            child {node {$u_{17}^4\ x_{17}$}}
                            child {node {$u_{18}^4\ x_{18}$}}}
                        }
                    child {node {$u_4^2$}
                        child {node {$u_7^3$}
                            child {node {$u_{19}^4\ x_{19}$}}
                            child {node {$u_{20}^4\ x_{20}$}}
                            child {node {$u_{21}^4\ x_{21}$}}}
                        child {node {$u_8^3$}
                            child {node {$u_{22}^4\ x_{22}$}}
                            child {node {$u_{23}^4\ x_{23}$}}
                            child {node {$u_{24}^4\ x_{24}$}}}}}
                child {node {$u_3^1$}
                    child {node {$u_5^2$}
                        child {node {$u_5^3$}
                            child {node {$u_{25}^4\ x_{25}$}}
                            child {node {$u_{26}^4\ x_{26}$}}
                            child {node {$u_{27}^4\ x_{27}$}}}
                        child {node {$u_6^3$}
                            child {node {$u_{28}^4\ x_{28}$}}
                            child {node {$u_{29}^4\ x_{29}$}}
                            child {node {$u_{30}^4\ x_{30}$}}}
                        }
                    child {node {$u_6^2$}
                        child {node {$u_7^3$}
                            child {node {$u_{31}^4\ x_{31}$}}
                            child {node {$u_{32}^4\ x_{32}$}}
                            child {node {$u_{33}^4\ x_{33}$}}}
                        child {node {$u_8^3$}
                            child {node {$u_{34}^4\ x_{34}$}}
                            child {node {$u_{35}^4\ x_{35}$}}
                            child {node {$u_{36}^4\ x_{36}$}}}}}
                child {node {$u_4^1$}
                    child {node {$u_7^2$}
                        child {node {$u_{13}^4\ x_{37}$}}
                        child {node {$u_{14}^4\ x_{38}$}}
                        child {node {$u_{15}^4\ x_{39}$}}}
                    child {node {$u_8^2$}
                        child {node {$u_{16}^4\ x_{40}$}}
                        child {node {$u_{17}^4\ x_{41}$}}
                        child {node {$u_{18}^4\ x_{42}$}}}};
    \end{tikzpicture}
    \caption{Tree structure for Cox regression.}
    \label{fig:simu_tree_cox}
\end{figure}
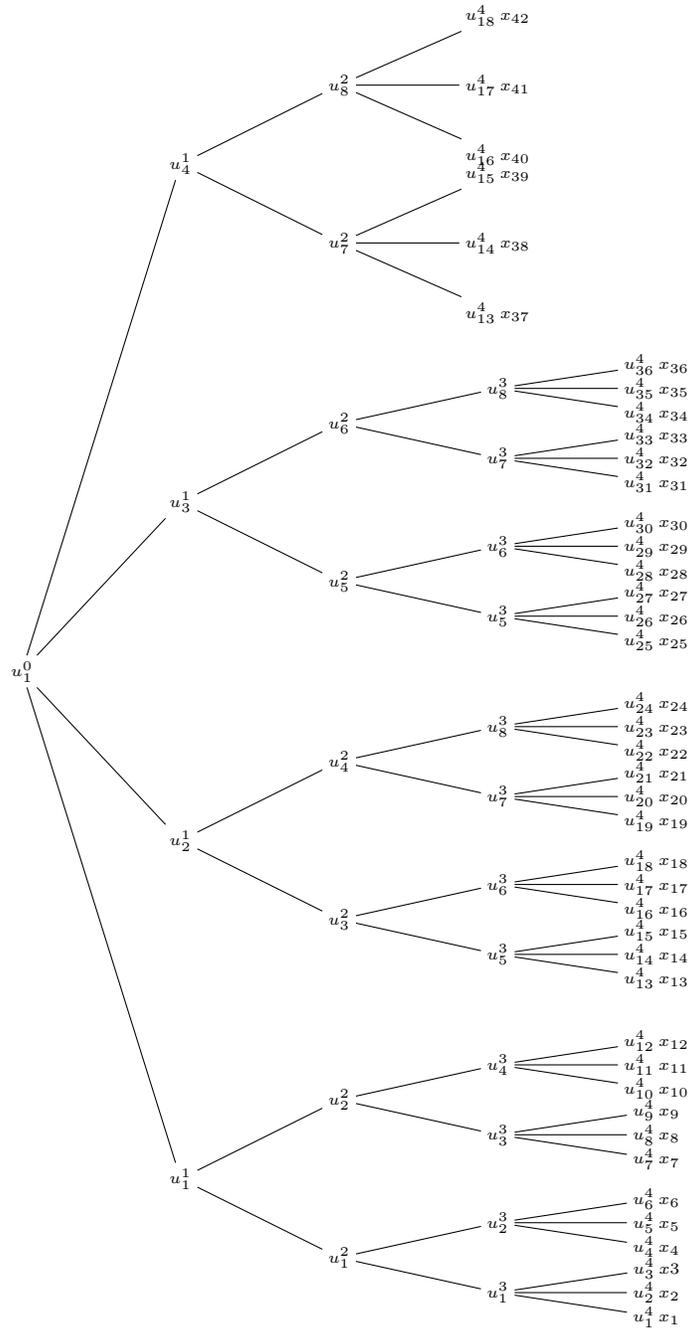

Figure~\ref{fig:simu_tree_cox} gives the hierarchical structure for simulation studies under the Cox regression settings.

%\section*{Acknowledgments}

%The authors are grateful to the Editor, the Associate Editor, and the referee for their valuable comments and suggestions, which have led to significant improvement of the article.

%\input{appendix}

\end{document}